\pdfoutput=1

\documentclass[11pt]{article}

\usepackage[final]{acl}

\usepackage{times}
\usepackage{latexsym}

\usepackage[T1]{fontenc}

\usepackage[utf8]{inputenc}

\usepackage{microtype}

\usepackage{inconsolata}

\usepackage{graphicx}

\usepackage{booktabs}
\usepackage{multirow}
\usepackage{enumerate}
\usepackage{nicematrix}
\usepackage{arydshln}
\usepackage{xcolor}

\newcommand{\ul}[1]{{\underline{#1}}}

\title{Decoupling Content and Expression: Two-Dimensional Detection of AI-Generated Text}

\author{
    Guangsheng Bao\textsuperscript{\rm 1,2,\footnotemark[1]},
    Lihua Rong\textsuperscript{\rm 3,\footnotemark[1]},
    Yanbin Zhao\textsuperscript{\rm 4},
    Qiji Zhou\textsuperscript{\rm 2}, 
    and Yue Zhang\textsuperscript{\rm 2,\footnotemark[2]}
    \\
    \textsuperscript{1} Zhejiang University \hspace{15pt} 
    \textsuperscript{2} School of Engineering, Westlake University \\
    \textsuperscript{3} Zhejiang University of Technology \hspace{15pt} 
    \textsuperscript{4} Shanghai Polytechnic University \\    
    \texttt{ronglihua1981@zjut.edu.cn}  \hspace{15pt} \texttt{zhaoyb553@nenu.edu.cn} \\
        \texttt{\{baoguangsheng,zhouqiji,zhangyue\}@westlake.edu.cn} \\
}

\begin{document}
\maketitle

\renewcommand{\thefootnote}{\fnsymbol{footnote}}
\footnotetext[1]{Equal contribution. \footnotemark[2]Corresponding author.}
\renewcommand{\thefootnote}{\arabic{footnote}}

\begin{abstract}
The wide usage of LLMs raises critical requirements on detecting AI participation in texts. Existing studies investigate these detections in scattered contexts, leaving a systematic and unified approach unexplored. In this paper, we present \emph{HART}, a hierarchical framework of AI risk levels, each corresponding to a detection task. To address these tasks, we propose a novel \emph{2D Detection Method}, decoupling a text into content and language expression. Our findings show that content is resistant to surface-level changes, which can serve as a key feature for detection. Experiments demonstrate that 2D method significantly outperforms existing detectors, achieving an AUROC improvement from 0.705 to 0.849 for level-2 detection and from 0.807 to 0.886 for RAID. We release our data and code at \url{https://github.com/baoguangsheng/truth-mirror}.

\end{abstract}

\section{Introduction}
\begin{figure}[t]
    \centering
    \includegraphics[width=0.9\linewidth]{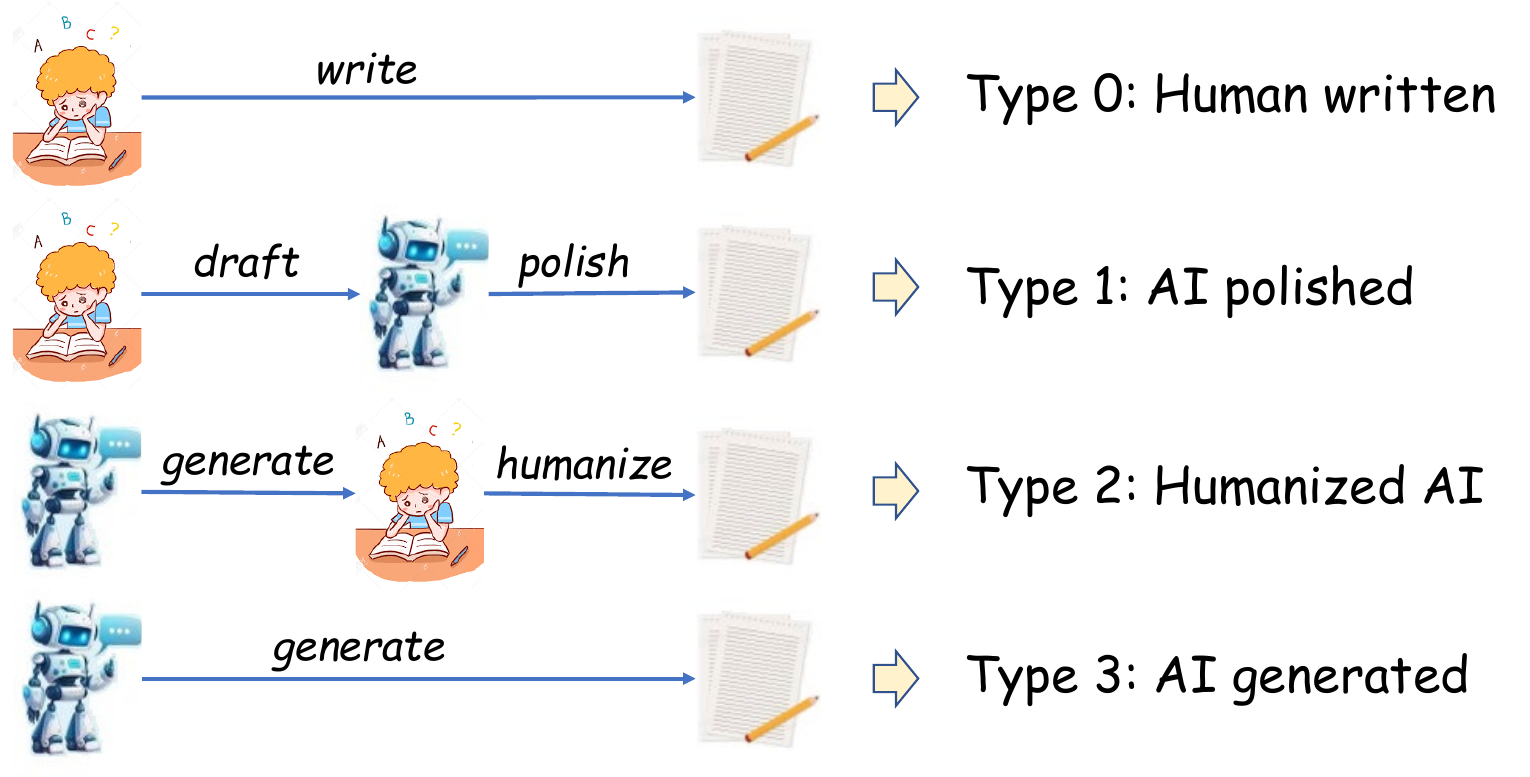}
    \caption{AI participation in text creation}
    \label{fig:llm-assistance}
\end{figure}

Large language models (LLMs) have shown strong text generation abilities, leading to the rise of AI-assisted text creation in news, academic, story, and advertising writing \cite{christian2023cnet, m2022exploring, yuan2022wordcraft, chen2023large}. The coauthorship between humans and machines has become the norm in the era of LLM \cite{lee2022coauthor,nguyen2024human,liang2024mapping}. However, we have different levels of tolerance for AI in different contexts. For example, in academic paper writing, conferences and journals usually accept papers polished using LLMs but reject papers fabricated by models. In writing class, teachers prefer the essays written completely by students, denying the usage of AI. These application scenarios require techniques to detect AI participation in text creation at varying levels, which can be categorized into four types as illustrated in Figure \ref{fig:llm-assistance}.

\begin{figure*}[t]
    \centering
    \includegraphics[trim={0pt 0pt 0pt 0pt},clip,width=1.0\linewidth]{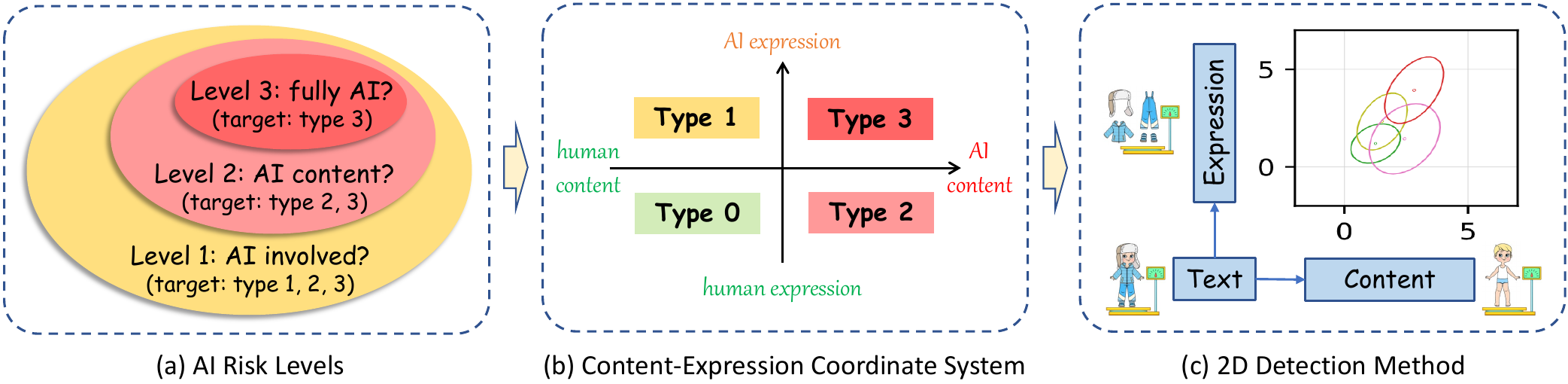}   
    \caption{The detection tasks across three risk levels address the four types of AI participation. We represent these types in a two-dimensional space, leading to a 2D detection approach. In this method, the detector performs a binary classification within the two-dimensional space for each detection task.}
    \label{fig:2d_detection2}
\end{figure*}

Prior research explores detection of AI-generated text across different contexts. Early studies concentrate on identifying fully AI-generated text \cite{gehrmann2019gltr,ippolito2020automatic,mitchell2023detectgpt}, while later studies address challenges like paraphrasing and adversarial attacks \cite{krishna2024paraphrasing,he2024mgtbench,dugan2024raid,wu2024detectrl}. Recently, the focus shifts toward identifying mixed human-AI content \cite{wang2024m4,richburg2024automatic,zhang2024llm,abassy2024llm}. However, these studies, tailored to specific contexts and detector designs, lack a systematic framework capable of addressing all levels of AI participation in a unified manner.

In this paper, we introduce \textbf{HART} (\textbf{H}ierarchical \textbf{A}I \textbf{R}isk in \textbf{T}ext Creation), a comprehensive framework of AI risk levels that targets the four types of AI participation, as depicted in Figure \ref{fig:2d_detection2}(a). Each risk level corresponds to a detection task, where a binary classifier is required. To systematically tackle these tasks, we propose decoupling the content and language expression of a text, as illustrated in Figure \ref{fig:2d_detection2}(b).
We map the four types of AI participation onto the four quadrants of a two-dimensional coordinate system, where type 2 and type 3 (AI content) are marked as high risk (in red) due to the potential for misinformation, bias, or harmful content, while type 1 (AI expression) is considered low risk (in yellow) as it primarily affects readers' experience.
Based on the two-dimensional view, we propose a novel \textbf{2D Detection Method} that decomposes the problem into two sub-problems: detecting AI content and detecting AI expression. Each corresponds to a distinct textual feature, mapped to a scalar metric, as Figure \ref{fig:2d_detection2}(c) illustrates.

We hypothesize that content is the essential feature in distinguishing AI-generated texts from human-written ones. This is because content is relatively stable and less affected by superficial text changes. For example, the content ``{\it the sun rises in the east}'' can be expressed in different styles -- academic as ``{\it the sun appears to ascent from the eastern horizon}'', news as ``{\it the sun rose in the east this morning, marking another predictable beginning to the day for residents in the region}'', and poetry as ``{\it in dawn's gentle embrace, the golden orb doth rise, from eastern realms it paints the morning skies}''. The core ideas of the content remain the same no matter whether words, grammar, style, and tone are altered.

To test this hypothesis, we investigate two fundamental research questions. 
\begin{enumerate}[{\it Q1:}]
\it
    \item How can a text’s content and expression be effectively decoupled?
    \item How can AI content and AI expression be reliably detected?
\end{enumerate}
For Q1, we explore two prototyping approaches: \emph{extraction}, which isolates main ideas or key expressions to represent content, and \emph{neutralization}, which simplifies the text by removing unique stylistic elements or ideas. For Q2, we assess existing detection models on these content-driven and expression-driven representations, finding that current metric-based detectors can indeed be adapted to identify both AI-generated content and AI-modified expressions.

Experimental results reveal that existing detectors struggle with AI-risk detection tasks due to their high sensitivity to surface-level text changes. In contrast, leveraging content-based features proves more robust, outperforming traditional detectors across multiple domains in the HART and RAID datasets. Further improvements are observed when content and expression features are integrated; the 2D framework boosts the best AUROC for the level-2 detection task from 0.711 to 0.855 and TPR5\% from 47\% to 59\%; and it enhances the best AUROC on RAID from 0.807 to 0.886.

To our knowledge, this is the first work to tackle the detection problem by focusing on \emph{content} as a key feature, demonstrating its importance in distinguishing AI-generated texts and effectively mitigating diverse \emph{attacks} on detection systems.

\section{Related Work}

\paragraph{AI-Assisted Text Creation.}
LLMs have made significant progress in the area of creative assistance \cite{zhao2023more,lund2023chatgpt,wasi2024llms}. These models can generate coherent, natural text, offer a variety of writing styles and expressions, and are adapted for various writing tasks such as scientific technology \cite{gero2022sparks,salimi2023large,lund2023chatgpt}, storytelling \cite{yuan2022wordcraft,zhao2023more,wang2024weaver,wang2024storyverse}, and news media \cite{cheng2024snil}. On the one hand, LLMs can help creators improve their writing efficiency, and on the other hand, they can enhance the quality of their writing.
In this paper, we categorize the ways in which LLMs assist in creating text content into four types and propose detection tasks to cover them.


\paragraph{AI-Generated Text Detection.}
The tasks we propose are related to existing AI-generated text detection tasks \cite{wu2023survey,yang2023survey}, where existing tasks do not consider AI participation levels in text creation. It is also related to existing refined detection tasks \cite{zhang2024llm,richburg2024automatic}, where these tasks identify operations applied to text by LLMs or humans. In contrast, we focus on AI risk levels instead of specific operations. 


Existing detectors consist of three types of technology. The first is supervised classifiers\cite{solaiman2019release,ippolito2020automatic,fagni2021tweepfake,hu2023radar,yan2023detection,li2024mage,verma2024ghostbuster}, which train a binary classifier based on a large collection of machine-generated and human-written text. The second is zero-shot classifiers, including white-box methods \cite{gehrmann2019gltr,su2023detectllm,bao2024fast,xu2024detecting,hans2024spotting} and black-box methods \cite{mitchell2023detectgpt,yang2023dna,bhattacharjee2024fighting,bao2025glimpse}. These technologies usually use pre-trained language models to extract detection metrics. The third is text watermarking technology \cite{kirchenbauer2023watermark,zhao2023protecting,christ2024undetectable, zhao2024permute,zhao2024sok}, which identifies machine-generated text by embedding easy-to-detect markers or patterns.

These techniques are effective in detecting purely machine-generated texts, but may not be robust to various attacks \cite{gao2018black,dyrmishi2023humans,krishna2024paraphrasing,he2024mgtbench,dugan2024raid,wu2024detectrl,wang2024stumbling}. At the same time, various commercial AI systems are published to serve `humanizing' ability, bypass existing detectors. To address these challenges, we propose the 2D detection framework as an effective candidate to defend against attacks.

\paragraph{Decoupling of Content and Expression.}
The idea of decoupling content and expression is related to existing studies on the disentanglement of semantics and syntax. These studies mainly focus on the disentanglement at the sentence level and discuss about it in different contexts, such as recognition science \cite{caucheteux2021disentangling,moro2001syntax}, sentence representation \cite{chen2019multi}, sentence comprehension \cite{dapretto1999form}, and sentence generation \cite{bao2019generating}. They generally represent semantics and syntax in separate neural vectors and train a neural network with specific structure or training objective to obtain disentangled vectors. 

However, our decoupling of content and expression differs from these early studies in three aspects. First, we focus on discourse level instead of sentence level, where the texts are longer and more complex. Second, we represent content and expression still in texts instead of neural vectors, which provides us with a convenience for understanding and explaining. Finally, we decouple them using zero-shot prompting techniques instead of training a model, which simplifies the usage.

\section{Task and Benchmark}
\label{sec:tasks}

\subsection{Task Definition}
We consider AI risks in the dimensions of content and expression, categorizing AI risks into three levels and defining detection tasks accordingly as Figure \ref{fig:2d_detection2}(a) shows.

\paragraph{Level-1 Detection:}
It targets types 1, 2, and 3, covering all texts in which their creation involves AI techniques. This task is suitable for strict situations where AI assistance is forbidden.

\paragraph{Level-2 Detection:}
It targets type 2 and 3, covering all texts whose contents are generated by AI. These texts may contain fabricated content that may deliver wrong, biased, or dangerous information. This task suites for common situations where AI content may cause risks. Existing AI-generated text detection tasks can be seen as level-2 detection.
 

\paragraph{Level-3 Detection:}
It targets type 3 only, where texts are generated by LLMs from scratch. This task suites for loose situations, where AI content is allowed, but readers' experience matters.
Early research in pure AI-generated text detection can be seen as level-3 detection. 

AI-assisted text creation in real scenarios is complex, where it is likely that human and AI participate iteratively. In this case, it is hard to define the risk levels. However, we could use the definition of level-1 to 3 as a lens to analyze the texts.

\subsection{Benchmark Dataset}

We create the benchmark dataset \emph{HART} for AI risk detection following a strict construction process and thorough quality assurance. Detailed statistics are provided in Table \ref{tab:datasets}, while information on model and parameter coverage can be found in Appendix \ref{app:data_coverage}.

\subsubsection{Data Construction}

To begin, we gather human-written texts from diverse sources, creating type-0 samples. Next, we refine these texts to produce type-1 samples, which preserve the original meanings but use different expressions. Using the titles or prompts of the human texts, we generate AI-written content, resulting in type-3 samples. We then adapt the AI-generated texts to create type-2 samples, which retain AI-generated content but are expressed in a more human-like manner. As a result, we obtain an equal distribution of samples across all four types.

\paragraph{Human Texts Collection.}
We consider the most common domains explored in AI-generated text detection research. Specifically, we utilize \emph{student essays} from Automated Student Assessment Prize (ASAP) 2.0 dataset \cite{scott2024asap}, \emph{paper introductions} sourced from arXiv \cite{arxiv2024}, \emph{story writings} taken from WritingPrompts \cite{fan2018hierarchical}, and \emph{news articles} obtained from Common Crawl \cite{Hamborg2017}, as detailed in Appendix \ref{app:datasets}.
The news articles are collected in five different languages. For every domain and language, we randomly sample 1000 examples, dividing them equally into development and test sets.

\paragraph{Automatic Refinement.}
LLMs are commonly employed to improve the expression of human-written drafts. We focus on two refinement methods: polishing and restructuring. \emph{Polishing} aims to enhance the readability and coherence of the text, typically adjusting language at the word and sentence levels. \emph{Restructuring}, on the other hand, focuses on improving the logical flow by reorganizing content, which demands a deeper grasp of the main ideas and the text's purpose. These refinement approaches are applied using the prompts outlined in Appendix \ref{app:refinement}.

\begin{table}[t]
    \centering\small
    \begin{tabular}{@{}llccc@{}}
        \toprule
        \bf Domain & \bf Language & \bf Length & \bf Dev & \bf Test \\
        \midrule
        Student Essay & English & 241 words & 2K & 2K \\
        ArXiv Intro & English & 410 words & 2K & 2K \\
        Creative Writing & English & 345 words & 2K & 2K \\
        CC News & English & 148 words & 2K & 2K \\
        CC News & Chinese & 590 chars & 2K & 2K \\
        CC News & French & 258 words & 2K & 2K \\
        CC News & Spanish & 285 words & 2K & 2K \\
        CC News & Arabic & 152 words & 2K & 2K \\
        \bottomrule
    \end{tabular}
    \caption{Domains and languages covered by HART.}
    \label{tab:datasets}
\end{table}

\paragraph{Machine Texts Generation.}
We create AI-generated texts using titles or prompts derived from human-written content. For instance, in the case of student essays, we instruct LLMs with a prompt such as: ``{\it Write a student essay (no title) in \{nwords\} words (split into \{nparagraphs\} paragraphs) based on the given title: \{title\}}''. To ensure that the generated texts closely match the average length of human-written texts, we specify the same number of words (or characters for Chinese) and paragraphs in the prompt. The detailed prompts for all domains can be found in Appendix \ref{app:datasets}.

\paragraph{Humanizing.}
AI-generated texts can be humanized to enhance their expressive quality. This can be achieved through human editing, the use of external tools, and two automated approaches: diversifying and mimicking. \emph{Diversifying} involves increasing the linguistic variety of AI-generated content, resembling the paraphrasing technique that enhances lexical and grammatical diversity \cite{krishna2024paraphrasing}. \emph{Mimicking}, on the other hand, prompts LLMs to emulate a human-written reference text in order to rewrite AI-produced content. These two strategies are applied using simple prompts, as detailed in Appendix \ref{app:humanizing}, while human editing and the use of external tools are discussed in Appendices \ref{app:human_editing} and \ref{app:external_tools}, respectively.

\subsection{Quality Assurance}
\label{app:quality_assure}
We evaluate the length of each generation from the LLM output, and if it is significantly longer (more than twice the original length) or shorter (less than half the original length), we prompt the LLM to generate the text again. Additionally, we monitor for issues like repetition or nonsensical responses and address them by re-generating the text. After processing the data, we truncate the texts to ensure the length distributions are consistent across types 0, 1, 2, and 3. As a final quality check, we randomly select 100 samples per domain for manual review, achieving an average pass rate of 99.5\%. In terms of costs, the data construction process involves approximately \$2,000 for LLM API usage, \$2,000 for human editing, \$500 for AI tools, and \$500 for the manual review step.

\begin{table}[t]
    \centering\scriptsize
    \begin{NiceTabular}{p{0.42\linewidth}p{0.48\linewidth}}
        \toprule
        \multicolumn{2}{p{0.95\linewidth}}{\textbf{Original Text ($T$)}: {\it Jane was walking through the park when she spotted a small puppy wandering aimlessly. She picked it up, read the collar tag, and took it to the address. The owner was grateful to have the puppy back.}} \\
        \midrule
        \textbf{Prompt 1:} {\it Outline the main points of the text to get a clear and concise picture of the content.} & \textbf{Prompt 1}: {\it Identify and list the representative language expressions used in the text.} \\
        \addlinespace[0.5em]
        \textbf{Content ($C_1$):} {\it 
        1. Jane was walking through the park.
        2. She noticed a small puppy wandering aimlessly.
        3. Jane found a collar tag on the puppy.
        4. She took the puppy to the address on the tag.
        5. The owner was grateful to Jane for returning the puppy.}  & \textbf{Expression ($E_1$):} {\it 
        1. "walking through the park" - indicates Jane's action and setting.
        2. "spotted a small puppy" - describes Jane noticing the puppy.
        3. "wandering aimlessly" - characterizes the puppy's behavior.
        4. "picked it up" - describes Jane's action of picking up the puppy.
        } \\
        \midrule
        \textbf{Prompt 2:} {\it Simplify the text to make it clear and concise while preserving its meaning.} & \textbf{Prompt 2:} {\it Replace the main points of the text with a generic topic while preserving the language expression.} \\
        \addlinespace[0.5em]
        \textbf{Content ($C_2$):} {\it Jane found a puppy in the park and returned it to its grateful owner after reading the collar tag.} & \textbf{Expression ($E_2$):} {\it Alex was strolling through the garden when they noticed a tiny kitten meandering without direction. They scooped it up, checked the collar tag, and brought it to the listed location. The caretaker was thankful to have the kitten returned.} \\
        \bottomrule
    \end{NiceTabular}
    \caption{Decoupling content and language using extraction and neutralization prompts.}
    \label{tab:decouple_content_language}
\end{table}

\section{2D Detection Method}
\label{sec:method}

We prototype to test our hypothesis using a simple zero-shot prompting technique.

\subsection{Decoupling Content and Expression}

Achieving a ``perfect decoupling'' of content and expression means presenting core ideas or meaning (content) in a way that is entirely independent from stylistic or linguistic expression.  
In this study, we propose a prototyping method by extracting content of a text and describing it in simple language to create a representation of the content, and discarding content of a text but keeping its language style and tone to create a representation of its expression.

Specifically, as Table \ref{tab:decouple_content_language} shows, we investigate two decoupling methods: extraction and neutralization. The extraction method produces a brief outline of the main ideas and a list of representative expressions of a text. Although the outline and list produced by the extraction method are relatively short and empirically effectual, they lose significant amount of details of the text. The neutralization method mitigates this issue. It reserves more details about the content and expression with longer text descriptions, which is empirically better for detection tasks.

\subsection{Detection of AI Content and Expression}

Intuitively, language models produce less surprising text than humans because the models are trained to minimize the empirical risk on human-written texts, which encourages the model to generate common patterns in the training data. Thus, AI-generated texts generally tend to have lower perplexity than human-written texts and can be detected by perplexity-based detectors. However, perplexity-based detectors are easily deceived by altered expressions because perplexity itself cannot distinguish a surprising content from a surprising expression. 

By decoupling content and expression, we can measure the surprisingness of them separately. Thus, many existing metric-based detectors can be used to detect AI content and expression. Take Fast-Detect as an example. As Figure \ref{fig:2d_detection2}(c) shows, its metric -- conditional probability curvature -- can be used to map the textual features into scalars, resulting in a two-dimensional distribution of texts for the four types. We also tried trained detectors such as RADAR, but failed to obtain an improvement in the AI content detection task. These detectors may need further training to handle the textual features. In our experiments, we empirically choose Fast-Detect and Binoculars as the representatives.

\begin{table*}[t]
    \centering\scriptsize
    \begin{tabular}{@{\hspace{2pt}}l@{\hspace{2pt}}|c@{\hspace{8pt}}c@{\hspace{5pt}}c@{\hspace{5pt}}c@{\hspace{2pt}}|c@{\hspace{8pt}}c@{\hspace{5pt}}c@{\hspace{5pt}}c@{\hspace{2pt}}|c@{\hspace{8pt}}c@{\hspace{5pt}}c@{\hspace{5pt}}c@{\hspace{2pt}}}
        \toprule
        \multirow{2}{*}{\bf Detector} & \multicolumn{4}{c}{\bf \colorbox{red!60}{Level-3} Detection Task} & \multicolumn{4}{c}{\bf \colorbox{red!40}{Level-2} Detection Task} & \multicolumn{4}{c}{\bf \colorbox{yellow!60}{Level-1} Detection Task} \\
         & \bf Essay & \bf ArXiv & \bf Writing & \bf ALL (TPR5\%) & \bf Essay & \bf ArXiv & \bf Writing & \bf ALL (TPR5\%) & \bf Essay & \bf ArXiv & \bf Writing & \bf ALL (TPR5\%) \\
        \midrule
        RoBERTa(ChatGPT) & 0.636 & 0.796 & 0.653 & 0.662 (16\%) & 0.435 & 0.687 & 0.498 & 0.502 (8\%) & 0.471 & \bf 0.955 & 0.606 & 0.566 (9\%) \\
        RADAR & 0.692 & 0.849 & 0.647 & 0.728 (14\%) & 0.566 & 0.814 & 0.630 & 0.687 (10\%) & 0.705 & \ul{0.857} & 0.700 & 0.758 (20\%) \\
        Log-Perplexity & 0.868 & 0.850 & 0.811 & 0.799 (33\%) & 0.364 & 0.485 & 0.438 & 0.473 (11\%) & 0.769 & 0.530 & 0.625 & 0.576 (6\%) \\
        Log-Rank & 0.867 & 0.874 & 0.813 & 0.814 (39\%) & 0.380 & 0.460 & 0.441 & 0.465 (11\%) & 0.739 & 0.542 & 0.611 & 0.573 (8\%) \\
        LRR & 0.835 & \bf 0.909 & 0.797 & 0.840 (50\%) & 0.560 & 0.616 & 0.551 & 0.573 (25\%) & 0.616 & 0.576 & 0.558 & 0.568 (19\%) \\
        Glimpse & \bf 0.929 & 0.869 & 0.819 & 0.849 (58\%) & 0.754 & 0.737 & 0.625 & 0.676 (30\%) & 0.878 & 0.719 & 0.618 & 0.688 (22\%) \\
        \hdashline
        Fast-Detect & 0.883 & 0.877 & 0.840 & 0.862 (60\%) & 0.734 & 0.718 & 0.692 & 0.711 (\ul{47\%}) & 0.877 & 0.769 & 0.740 & 0.778 (55\%) \\
        $C_2$ (Fast-Detect) & 0.734 & 0.787 & 0.765 & 0.738 (18\%) & 0.778 & 0.862 & 0.819 & 0.798 (42\%) & 0.712 & 0.779 & 0.742 & 0.730 (33\%) \\
        $C_2$-$T$ (Fast-Detect) & 0.864 & 0.896 & \ul{0.890} & \ul{0.876} (\ul{61\%}) & \bf 0.785 & \bf 0.915 & \ul{0.890} & \bf 0.855 (59\%) & \bf 0.907 & 0.849 & \bf 0.836 & \bf 0.843 (59\%) \\
        \hdashline
        Binoculars & \ul{0.897} & 0.882 & 0.847 & 0.870 ({\bf 62\%}) & 0.735 & 0.715 & 0.693 & 0.711 (44\%) & 0.879 & 0.769 & 0.740 & 0.780 (55\%) \\
        $C_2$ (Binoculars) & 0.736 & 0.789 & 0.770 & 0.737 (17\%) & \ul{0.781} & 0.856 & 0.822 & 0.791 (35\%) & 0.701 & 0.761 & 0.743 & 0.716 (25\%) \\
        $C_2$-$T$ (Binoculars) & 0.854 & \ul{0.904} & \bf 0.905 & {\bf 0.883} (\ul{61\%}) & 0.746 & \ul{0.913} & \bf 0.895 & \ul{0.848} (32\%) & \ul{0.900} & 0.840 & \ul{0.828} & \ul{0.838} (\ul{58\%}) \\
        \bottomrule
    \end{tabular}
    \caption{Results on AI risk detection, evaluated on HART. The best AUROCs and TPR5\% are marked in \textbf{bold} and second in \ul{underline}. The column `{\it ALL}' denotes a mixture of domains including Essay, arXiv, Writing, and News in English.}
    \label{tab:results_detect_level}
\end{table*}

\section{Experiments}

We confirm our hypothesis and demonstrate that combining content and expression features provides us with a stronger detection ability to AI risks in Section \ref{sec:detect_risk_level} and resilience to various attacks in Section \ref{sec:detect_ai_text}.

\subsection{Experimental Settings}

\paragraph{Detectors.}
We mainly focus on \emph{metric-based detectors}, which generally leverage existing pre-trained LLMs to compute a metric as an indicator of AI-generated text. We take \emph{log-perplexity}, \emph{log-rank}, \emph{LRR} \cite{su2023detectllm}, \emph{Fast-Detect} \cite{bao2024fast}, \emph{Binoculars} \cite{hans2024spotting}, and \emph{Glimpse} \cite{bao2025glimpse} as representatives, as described in Appendix \ref{app:detectors}. For fair comparison, we unify the scoring models to falcon-7B or falcon-7B-instruct (except for Glimpse), where we find that these models perform significantly better than smaller models such as gpt-neo-2.7B.

We also consider \emph{trained detectors}, such as RADAR \cite{hu2023radar} and RoBERTa (ChatGPT) \cite{guo2023hc3}. However, these detectors cannot detect extracted textual features without further training. Thus, we just list them for reference.

\paragraph{Metrics.}
We study the detection problem in various application scenarios, where the tolerance for false positive rate is unknown. Consequently, we use \emph{AUROC} (area under the receiver operating characteristic curve) as the major metric to measure the quality of the classifiers. We also report \emph{F1} and \emph{TPR5\%} (true positive rate at a false positive rate of 5\%) for reference.

\subsection{Results on Multi-Level AI Risk Detection}
\label{sec:detect_risk_level}

We compare existing detectors and 2D methods in HART as Table \ref{tab:results_detect_level} shows, achieving the following findings.

\paragraph{Finding 1: Existing detectors are good at level-3 detection but poor at level-2/1 detections.}

Existing detectors generally perform the best on the level-3 detection task, where Binoculars reaches an overall AUROC of 0.870 and TPR5\% of 62\%. These scores are significantly higher than those on level-2 and 1 detection tasks, suggesting that existing detectors may best suit pure AI-generated texts. This is potentially because existing detectors mainly measure texts along expression dimension, thus being sensitive to changes in language expressions.

\paragraph{Finding 2: The content feature is resilient to changes in language expression, resulting in better level-2 and 1 detection performance.}

When we compare 2D ($C_2$-$T$) methods with existing detectors, we find that although 2D methods perform at the same level as existing detectors on the level-3 detection task, they outperform existing detectors by a large margin on level-2 and 1 detection tasks. It increases overall AUROC from 0.704 to 0.849 on level-2 task and from 0.767 to 0.844 on level-1 task using Fast-Detect. Similarly, TPR5\% increases by 12\% and 4\%, respectively, on the two tasks. These results demonstrate the effectiveness of the 2D detection framework.

\begin{figure}[t]
    \centering
    \includegraphics[trim={0pt 0pt 0pt 0pt},clip,width=1.0\linewidth]{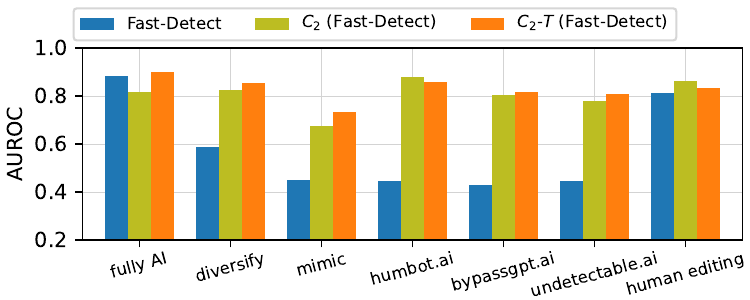}   
    \caption{Comparison on their ability to detect AI-generated texts, where `{\it xxx.ai}' are external humanizing tools.}
    \label{fig:level2_analysis_barchart}
\end{figure}

We further look into each type of level-2 detection data, as Figure \ref{fig:level2_analysis_barchart} shows.  The content feature plays a key role in the detection of humanized AI-generated texts, significantly outperforming the baseline. The results also confirm our hypothesis that content is the essential feature for identifying AI-generated texts.

\paragraph{Finding 3: The content feature is effective across languages.}
We evaluate the detectors across five languages in Appendix \ref{app:multilingual}. 2D detectors outperform the baselines on all three tasks, where the improvements on the level-2 and 1 tasks are especially significant. It is worth noting that Glimpse using gpt-3.5-turbo achieves the best overall results across the languages, possibly because of the stronger multilingual ability of the model.

\begin{figure}[t]
    \centering
    \includegraphics[trim={2pt 10pt 2pt 2pt},clip,width=1.0\linewidth]{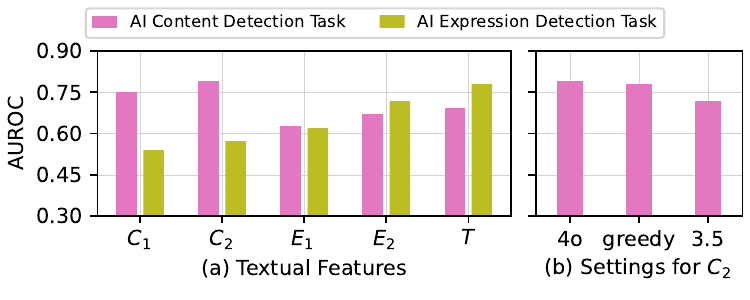}   
    \caption{Content and expression features evaluated on AI detection tasks using conditional probability curvature as the feature metric.}
    \label{fig:auroc_content_language}
\end{figure}

\begin{table*}[t]
    \centering\scriptsize
    \begin{tabular}{l|cccccccc|ccc}
        \toprule
        \bf Detector & \bf News & \bf Books & \bf Wiki & \bf Abstracts & \bf Reddit & \bf Recipes & \bf Poetry & \bf Reviews  & \bf ALL AUROC & \bf F1 & \bf TPR5\% \\
        \midrule
        RoBERTa-base & 0.588 & 0.622 & 0.582 & 0.643 & 0.673 & 0.500 & 0.638 & 0.710 & 0.614 & 67\% & 24\% \\
        RADAR & 0.884 & 0.912 & 0.842 & 0.842 & \bf 0.870 & 0.818 & 0.780 & 0.782 & 0.828 & 77\% & 42\% \\
        \hdashline
        Log-Perplexity & 0.644 & 0.725 & 0.701 & 0.680 & 0.725 & 0.627 & 0.706 & 0.698 & 0.663 & 66\% & 12\% \\
        Log-Rank & 0.666 & 0.745 & 0.719 & 0.701 & 0.735 & 0.645 & 0.725 & 0.716 & 0.681 & 67\% & 14\% \\
        LRR & 0.750 & 0.816 & 0.804 & 0.771 & 0.779 & 0.669 & 0.776 & 0.773 & 0.746 & 70\% & 34\% \\
        Glimpse & 0.712 & 0.758 & 0.589 & 0.787 & 0.742 & 0.670 & 0.756 & 0.728 & 0.715 & 67\% & 39\% \\
        Fast-Detect & 0.761 & 0.845 & 0.803 & 0.821 & 0.794 & 0.749 & 0.818 & 0.810 & 0.800 & 76\% & 54\% \\
        \hdashline
        Binoculars & 0.768 & 0.850 & 0.804 & 0.826 & 0.811 & 0.759 & 0.826 & 0.812 & 0.807 & 77\% & 58\% \\
        $C_2$ (Binoculars) & 0.783 & 0.888 & 0.808 & 0.799 & 0.778 & 0.726 & 0.777 & 0.762 & 0.774 & 72\% & 46\% \\
        $C_2$-$T$ (Binoculars) & \bf 0.901 & \bf 0.924 & \bf 0.861 & \bf 0.900 & 0.869 & \bf 0.878 & \bf 0.889 & \bf 0.869 & \bf 0.886 & \bf 82\% & \bf 68\% \\
        \bottomrule
    \end{tabular}
    \caption{Results on AI-generated text detection, evaluated on RAID. The highest AUROCs are marked in \textbf{bold}. $C_2$, $E_2$, and $T$ are textual features used for detection, which are illustrated in Table \ref{tab:decouple_content_language}.}
    \label{tab:results_detect_text}
\end{table*}

\subsubsection{Ablation Study}
\label{sec:detect_content_language}

\paragraph{Textual Features.}
The quality of \emph{extracted textual features} for content and expression is critical for detection tasks. We first evaluate the candidate features on AI content detection and AI expression detection tasks, each with 1000 pairs of samples from the HART dataset. We use conditional probability curvature as a metric to map textual features to scalar values. As Figure \ref{fig:auroc_content_language}(a) shows, neutralization generally outperforms extraction approach for both content and expression representations. The content feature $C_2$ achieves the best performance in the AI content detection task, while the original text $T$ achieves the best performance in the AI expression detection task, surpassing the expression features $E_1$ and $E_2$ with a significant margin. Thus, we choose $C_2$ as the content feature and $T$ as the expression feature for our 2D detectors.

\paragraph{Model and Parameters.}
The \emph{language model} and \emph{decoding strategy} can also affect the quality of the extracted content. We ablate them as Figure \ref{fig:auroc_content_language}(b).
Compared to the default setting of gpt-4o with random sampling, greedy decoding slightly decreases the AUROC. However, we find that greedy decoding further improves TPR5\% by about 4\% on the AI content detection task, which may be because deterministic decoding produces more stable texts. Changing the model to gpt-3.5-turbo causes a significant drop in AUROC, suggesting that a strong LLM is a prerequisite to extract effective content features. In our experiments, we use gpt-4o with random sampling.

\subsubsection{Analysis of Data Distribution}

\paragraph{What is the impact of source model and decoding parameters to generated texts?}
Various factors influence the distribution of AI-generated texts (type-3 texts), as described in Appendix \ref{app:analysis_params}. Among the source models, gpt-4o demonstrates the most diverse generations and is significantly closer to the origin of the coordinate system, suggesting its stronger ability to produce human-like texts. The decoding temperature also affects the distribution, but the differences are not significant. Similarly, larger top-$p$ and presence penalty produce more diverse texts but the differences are marginal. In contrast, the frequency penalty shows a strong impact on generated texts, where a larger penalty produces more human-like texts.

\paragraph{What has been changed by refinement and humanizing?}

As described in Appendix \ref{app:analysis_refinement_humanizing}, refinement and humanizing significantly alter the distribution, mainly along the expression dimension. The change bringing by humanizing is relatively bigger than refinement, where automatic humanizing shifts the distribution largely. Human editing alters the distribution not as significant as the automatic humanizing, which may be because that human annotators do not attack AI detectors purposely as the humanizing tools.

\subsubsection{Discussion}


Although content features are resilient to surface-level text changes, there is the possibility of developing attacks against the content of a text. However, we posit that attacking detectors at the content level is much harder than at the expression level. Meaningful and coherent content, unlike superficial language expression, requires deep understanding about the world, thus hard to be simulated by current language models. Additionally, a content-level attack may pay additional costs, such as reducing the logical coherence and readability of the generated content.

\begin{figure*}[t]
\begin{minipage}{0.34\linewidth}
    \centering
    \includegraphics[trim={0pt 0pt 0pt 0pt},clip,width=1.0\linewidth]{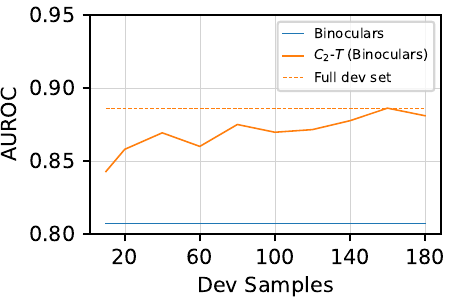}   
    \caption{Ablation on the number of dev samples required by 2D method.}
    \label{fig:raid_dev_samples_linechart}
\end{minipage}
\hfill\hspace{4pt}
\begin{minipage}{0.65\linewidth}
    \centering
    \includegraphics[trim={0pt 0pt 0pt 0pt},clip,width=1.0\linewidth]{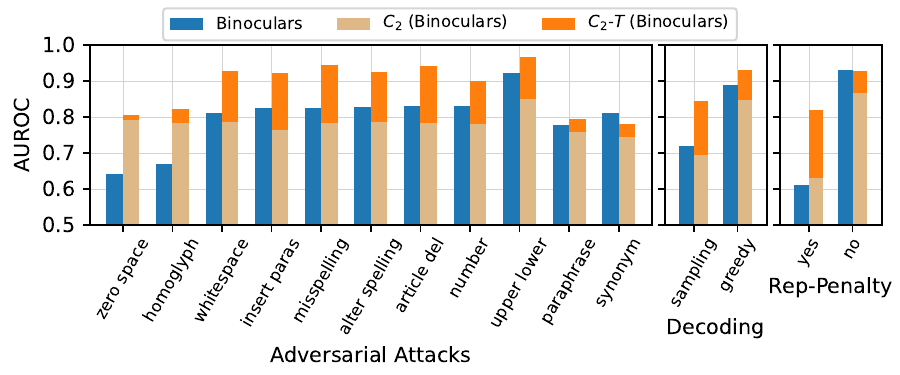}   
    \caption{Comparison on their ability to handle adversarial attacks, decoding strategies, and repetition penalty.}
    \label{fig:raid_analysis_barchart}
    
\end{minipage}
\end{figure*}

\subsection{Results on AI-Generated Text Detection}
\label{sec:detect_ai_text}

We evaluate 2D methods on existing detection tasks. Typically, we use the challenging RAID \cite{dugan2024raid} dataset, from which we sample 4K samples (250 pairs per domain) for testing and another 4K for development.

As Table \ref{tab:results_detect_text} shows, the columns AUROC, F1, and TPR@5\% are evaluated across all domains, where we find the best development threshold for calculating F1. As the AUROCs indicate, the content feature $C_2$ outperforms the baseline Binoculars on News, Books, and Wiki. When combining content and expression features, 2D ($C_2$-$T$) produces the best scores across all the domains. These results suggest the positive effect of content feature on the detection of AI-generated text.

Readers may wonder why the content feature does not outperform the baseline in all domains. We speculate that it is correlated to the genre and length of the text. Take poetry as an example. Poetry often relies on evocative language to convey emotions, themes, and ideas. The meaning of a poem can change depending on how it is expressed. Thus, the decoupling of content and expression loses significant information. Similar situations may happen with other free-style texts, such as Reviews and Reddit. (tbd)

\paragraph{Ablation on expression features.}
Empirically, $C_2$-$T$ outperforms $C_2$-$E_2$ on almost all domains, suggesting that the original text $T$ presents its language expression better than the extracted feature $E_2$. It confirms our finding in Section \ref{sec:detect_content_language} that the original text best represents its expression, while the extracted content feature $C_2$ best represents its content.

\paragraph{Ablation on the size of development set.}
2D methods need to fit a two-dimensional binary classifier, which requires additional samples. We show that such a classifier requires only a small number of samples because of its low dimensions. As Figure \ref{fig:raid_dev_samples_linechart} shows, 10 random samples are sufficient to outperform the baseline, resulting in a AUROC of 0.8428 in all domains. Empirically, 200 samples are sufficient to reach the full level of performance.

\paragraph{Analysis on attacks, decoding strategies, and repetition penalty.}
As Figure \ref{fig:raid_analysis_barchart} shows, the content feature outperforms the baseline on `{\it zero-width-space}', `{\it homoglyph}', and `{\it repetition-penalty}', demonstrating its effectiveness. When we combine content and expression features, we achieve significant improvements on all categories except synonym and non-repetition-penalty. The significant improvements on sampling and repetition-penalty suggest that the 2D method is typical beneficial for hard detection situations, given that sampling and repetition-penalty produce more nature texts which are harder for detection. These results suggest the advantage of the 2D method which is resistant to various attacks and decoding strategies.

\paragraph{Addressing the bias toward nonnative writers.}

Content representation is also resilient to nonnative English writers, where unique language expressions are reduced during content extraction. Consequently, using the content feature $C_2$ improves the AUROC from 0.4970 (Binoculars) to 0.5212 ($C_2$ with Binoculars). When we use the best threshold found on RAID development set, it improves F1 from 49\% (Binoculars) to 55\% ($C_2$ with Binoculars), demonstrating that the content feature reduces the bias toward nonnative writers.

\section{Conclusion}
We introduce a hierarchical framework for detection tasks, categorized into three levels of AI risk, which integrates prior research and established requirements. Our study explores 2D detection methods that leverage content as a key feature for identifying AI-generated text, demonstrating that content plays a critical role in addressing such detection challenges. Experimental results indicate that content features exhibit resilience to superficial textual modifications, making them a reliable tool for both emerging AI risk detection and traditional AI-generated text identification tasks. Furthermore, our proposed framework and benchmark dataset lay a strong foundation for advancing future research in this field.

\section*{Limitations}
The decoupling of content and expression may have various solutions, where prompting techniques may be the simplest, but not necessarily the most effective. There is the possibility to decouple content and expression in a more fundamental approach, with a specific design of a model or an algorithm.
On the other hand, detection of content and expression may require different methods given that they are at different levels of text. Therefore, a specific design for each may produce stronger detectors.

\section*{Ethical Considerations}
The dataset we use contains AI-generated texts, which could potentially be biased, offensive, or irresponsible. Although we filter them with automatic API provided by Azure OpenAI and check 10\% samples manually with high pass rate, there are still possibility of having unpleasant content in the released dataset, which may deserve a warning.


\bibliography{custom}

\begin{thebibliography}{67}
\providecommand{\natexlab}[1]{#1}

\bibitem[{Abassy et~al.(2024)Abassy, Elozeiri, Aziz, Ta, Tomar, Adhikari, Ahmed, Wang, Afzal, Xie et~al.}]{abassy2024llm}
Mervat Abassy, Kareem Elozeiri, Alexander Aziz, Minh Ta, Raj Tomar, Bimarsha Adhikari, Saad Ahmed, Yuxia Wang, Osama~Mohammed Afzal, Zhuohan Xie, et~al. 2024.
\newblock Llm-detectaive: a tool for fine-grained machine-generated text detection.
\newblock In \emph{Proceedings of the 2024 Conference on Empirical Methods in Natural Language Processing: System Demonstrations}, pages 336--343.

\bibitem[{arXiv(2024)}]{arxiv2024}
arXiv. 2024.
\newblock \href {https://arxiv.org/} {arxiv}.
\newblock \emph{Cornell University}.

\bibitem[{Banarescu et~al.(2013)Banarescu, Bonial, Cai, Georgescu, Griffitt, Hermjakob, Knight, Koehn, Palmer, and Schneider}]{banarescu2013abstract}
Laura Banarescu, Claire Bonial, Shu Cai, Madalina Georgescu, Kira Griffitt, Ulf Hermjakob, Kevin Knight, Philipp Koehn, Martha Palmer, and Nathan Schneider. 2013.
\newblock Abstract meaning representation for sembanking.
\newblock In \emph{Proceedings of the 7th linguistic annotation workshop and interoperability with discourse}, pages 178--186.

\bibitem[{Bao et~al.(2025)Bao, Zhao, He, and Zhang}]{bao2025glimpse}
Guangsheng Bao, Yanbin Zhao, Juncai He, and Yue Zhang. 2025.
\newblock Glimpse: Enabling white-box methods to use proprietary models for zero-shot llm-generated text detection.
\newblock \emph{The Thirteenth International Conference on Learning Representations}.

\bibitem[{Bao et~al.(2024)Bao, Zhao, Teng, Yang, and Zhang}]{bao2024fast}
Guangsheng Bao, Yanbin Zhao, Zhiyang Teng, Linyi Yang, and Yue Zhang. 2024.
\newblock Fast-detectgpt: Efficient zero-shot detection of machine-generated text via conditional probability curvature.
\newblock In \emph{The Twelfth International Conference on Learning Representations}.

\bibitem[{Bao et~al.(2019)Bao, Zhou, Huang, Li, Mou, Vechtomova, Dai, and Chen}]{bao2019generating}
Yu~Bao, Hao Zhou, Shujian Huang, Lei Li, Lili Mou, Olga Vechtomova, Xinyu Dai, and Jiajun Chen. 2019.
\newblock Generating sentences from disentangled syntactic and semantic spaces.
\newblock In \emph{Proceedings of the 57th Annual Meeting of the Association for Computational Linguistics}, pages 6008--6019.

\bibitem[{Bhattacharjee and Liu(2024)}]{bhattacharjee2024fighting}
Amrita Bhattacharjee and Huan Liu. 2024.
\newblock Fighting fire with fire: can chatgpt detect ai-generated text?
\newblock \emph{ACM SIGKDD Explorations Newsletter}, 25(2):14--21.

\bibitem[{Caucheteux et~al.(2021)Caucheteux, Gramfort, and King}]{caucheteux2021disentangling}
Charlotte Caucheteux, Alexandre Gramfort, and Jean-Remi King. 2021.
\newblock Disentangling syntax and semantics in the brain with deep networks.
\newblock In \emph{International conference on machine learning}, pages 1336--1348. PMLR.

\bibitem[{Chen et~al.(2019)Chen, Tang, Wiseman, and Gimpel}]{chen2019multi}
Mingda Chen, Qingming Tang, Sam Wiseman, and Kevin Gimpel. 2019.
\newblock A multi-task approach for disentangling syntax and semantics in sentence representations.
\newblock In \emph{Proceedings of the 2019 Conference of the North American Chapter of the Association for Computational Linguistics: Human Language Technologies, Volume 1 (Long and Short Papers)}, pages 2453--2464.

\bibitem[{Chen and Chan(2023)}]{chen2023large}
Zenan Chen and Jason Chan. 2023.
\newblock Large language model in creative work: The role of collaboration modality and user expertise.
\newblock \emph{Available at SSRN 4575598}.

\bibitem[{Cheng et~al.(2024)Cheng, Deng, Xie, Qiu, Xu, and Wu}]{cheng2024snil}
Liqi Cheng, Dazhen Deng, Xiao Xie, Rihong Qiu, Mingliang Xu, and Yingcai Wu. 2024.
\newblock Snil: Generating sports news from insights with large language models.
\newblock \emph{IEEE Transactions on Visualization and Computer Graphics}.

\bibitem[{Christ et~al.(2024)Christ, Gunn, and Zamir}]{christ2024undetectable}
Miranda Christ, Sam Gunn, and Or~Zamir. 2024.
\newblock Undetectable watermarks for language models.
\newblock In \emph{The Thirty Seventh Annual Conference on Learning Theory}, pages 1125--1139. PMLR.

\bibitem[{Christian(2023)}]{christian2023cnet}
Jon Christian. 2023.
\newblock Cnet secretly used ai on articles that didn’t disclose that fact, staff say.
\newblock \emph{Futurusm, January}.

\bibitem[{Crossley et~al.(2024)Crossley, Baffour, King, Burleigh, Reade, and Demkin}]{scott2024asap}
Scott Crossley, Perpetual Baffour, Jules King, Lauryn Burleigh, Walter Reade, and Maggie Demkin. 2024.
\newblock \href {https://www.kaggle.com/datasets/lburleigh/asap-2-0} {Asap 2.0: Automated student assessment prize}.
\newblock \emph{Kaggle}.

\bibitem[{Dapretto and Bookheimer(1999)}]{dapretto1999form}
Mirella Dapretto and Susan~Y Bookheimer. 1999.
\newblock Form and content: dissociating syntax and semantics in sentence comprehension.
\newblock \emph{Neuron}, 24(2):427--432.

\bibitem[{Dugan et~al.(2024)Dugan, Hwang, Trhlik, Ludan, Zhu, Xu, Ippolito, and Callison-Burch}]{dugan2024raid}
Liam Dugan, Alyssa Hwang, Filip Trhlik, Josh~Magnus Ludan, Andrew Zhu, Hainiu Xu, Daphne Ippolito, and Chris Callison-Burch. 2024.
\newblock Raid: A shared benchmark for robust evaluation of machine-generated text detectors.
\newblock \emph{arXiv preprint arXiv:2405.07940}.

\bibitem[{Dyrmishi et~al.(2023)Dyrmishi, GHAMIZI, and Cordy}]{dyrmishi2023humans}
Salijona Dyrmishi, Salah GHAMIZI, and Maxime Cordy. 2023.
\newblock How do humans perceive adversarial text? a reality check on the validity and naturalness of word-based adversarial attacks.
\newblock In \emph{The 61st Annual Meeting Of The Association For Computational Linguistics}.

\bibitem[{Fagni et~al.(2021)Fagni, Falchi, Gambini, Martella, and Tesconi}]{fagni2021tweepfake}
Tiziano Fagni, Fabrizio Falchi, Margherita Gambini, Antonio Martella, and Maurizio Tesconi. 2021.
\newblock Tweepfake: About detecting deepfake tweets.
\newblock \emph{Plos one}, 16(5):e0251415.

\bibitem[{Fan et~al.(2018)Fan, Lewis, and Dauphin}]{fan2018hierarchical}
Angela Fan, Mike Lewis, and Yann Dauphin. 2018.
\newblock Hierarchical neural story generation.
\newblock In \emph{Proceedings of the 56th Annual Meeting of the Association for Computational Linguistics (Volume 1: Long Papers)}. Association for Computational Linguistics.

\bibitem[{Fillmore et~al.(2006)}]{fillmore2006frame}
Charles~J Fillmore et~al. 2006.
\newblock Frame semantics.
\newblock \emph{Cognitive linguistics: Basic readings}, 34:373--400.

\bibitem[{Gao et~al.(2018)Gao, Lanchantin, Soffa, and Qi}]{gao2018black}
Ji~Gao, Jack Lanchantin, Mary~Lou Soffa, and Yanjun Qi. 2018.
\newblock Black-box generation of adversarial text sequences to evade deep learning classifiers.
\newblock In \emph{2018 IEEE Security and Privacy Workshops (SPW)}, pages 50--56. IEEE.

\bibitem[{Gehrmann et~al.(2019)Gehrmann, Strobelt, and Rush}]{gehrmann2019gltr}
Sebastian Gehrmann, Hendrik Strobelt, and Alexander~M Rush. 2019.
\newblock Gltr: Statistical detection and visualization of generated text.
\newblock In \emph{Proceedings of the 57th Annual Meeting of the Association for Computational Linguistics: System Demonstrations}, pages 111--116.

\bibitem[{Gero et~al.(2022)Gero, Liu, and Chilton}]{gero2022sparks}
Katy~Ilonka Gero, Vivian Liu, and Lydia Chilton. 2022.
\newblock Sparks: Inspiration for science writing using language models.
\newblock In \emph{Proceedings of the 2022 ACM Designing Interactive Systems Conference}, pages 1002--1019.

\bibitem[{Guo et~al.(2023)Guo, Zhang, Wang, Jiang, Nie, Ding, Yue, and Wu}]{guo2023hc3}
Biyang Guo, Xin Zhang, Ziyuan Wang, Minqi Jiang, Jinran Nie, Yuxuan Ding, Jianwei Yue, and Yupeng Wu. 2023.
\newblock How close is chatgpt to human experts? comparison corpus, evaluation, and detection.
\newblock \emph{arXiv preprint arxiv:2301.07597}.

\bibitem[{Hamborg et~al.(2017)Hamborg, Meuschke, Breitinger, and Gipp}]{Hamborg2017}
Felix Hamborg, Norman Meuschke, Corinna Breitinger, and Bela Gipp. 2017.
\newblock \href {https://doi.org/10.5281/zenodo.4120316} {news-please: A generic news crawler and extractor}.
\newblock In \emph{Proceedings of the 15th International Symposium of Information Science}, pages 218--223.

\bibitem[{Hans et~al.(2024)Hans, Schwarzschild, Cherepanova, Kazemi, Saha, Goldblum, Geiping, and Goldstein}]{hans2024spotting}
Abhimanyu Hans, Avi Schwarzschild, Valeriia Cherepanova, Hamid Kazemi, Aniruddha Saha, Micah Goldblum, Jonas Geiping, and Tom Goldstein. 2024.
\newblock Spotting llms with binoculars: Zero-shot detection of machine-generated text.
\newblock In \emph{Forty-first International Conference on Machine Learning}.

\bibitem[{He et~al.(2024)He, Shen, Chen, Backes, and Zhang}]{he2024mgtbench}
Xinlei He, Xinyue Shen, Zeyuan Chen, Michael Backes, and Yang Zhang. 2024.
\newblock Mgtbench: Benchmarking machine-generated text detection.
\newblock In \emph{Proceedings of the 2024 on ACM SIGSAC Conference on Computer and Communications Security}, pages 2251--2265.

\bibitem[{Hogan et~al.(2021)Hogan, Blomqvist, Cochez, d’Amato, Melo, Gutierrez, Kirrane, Gayo, Navigli, Neumaier et~al.}]{hogan2021knowledge}
Aidan Hogan, Eva Blomqvist, Michael Cochez, Claudia d’Amato, Gerard~De Melo, Claudio Gutierrez, Sabrina Kirrane, Jos{\'e} Emilio~Labra Gayo, Roberto Navigli, Sebastian Neumaier, et~al. 2021.
\newblock Knowledge graphs.
\newblock \emph{ACM Computing Surveys (Csur)}, 54(4):1--37.

\bibitem[{Hu et~al.(2023)Hu, Chen, and Ho}]{hu2023radar}
Xiaomeng Hu, Pin-Yu Chen, and Tsung-Yi Ho. 2023.
\newblock Radar: Robust ai-text detection via adversarial learning.
\newblock \emph{Advances in Neural Information Processing Systems}, 36:15077--15095.

\bibitem[{Ippolito et~al.(2020)Ippolito, Duckworth, Callison-Burch, and Eck}]{ippolito2020automatic}
Daphne Ippolito, Daniel Duckworth, Chris Callison-Burch, and Douglas Eck. 2020.
\newblock Automatic detection of generated text is easiest when humans are fooled.
\newblock In \emph{Proceedings of the 58th Annual Meeting of the Association for Computational Linguistics}, pages 1808--1822.

\bibitem[{Kirchenbauer et~al.(2023)Kirchenbauer, Geiping, Wen, Katz, Miers, and Goldstein}]{kirchenbauer2023watermark}
John Kirchenbauer, Jonas Geiping, Yuxin Wen, Jonathan Katz, Ian Miers, and Tom Goldstein. 2023.
\newblock A watermark for large language models.
\newblock In \emph{International Conference on Machine Learning}, pages 17061--17084. PMLR.

\bibitem[{Krishna et~al.(2024)Krishna, Song, Karpinska, Wieting, and Iyyer}]{krishna2024paraphrasing}
Kalpesh Krishna, Yixiao Song, Marzena Karpinska, John Wieting, and Mohit Iyyer. 2024.
\newblock Paraphrasing evades detectors of ai-generated text, but retrieval is an effective defense.
\newblock \emph{Advances in Neural Information Processing Systems}, 36.

\bibitem[{Lee et~al.(2022)Lee, Liang, and Yang}]{lee2022coauthor}
Mina Lee, Percy Liang, and Qian Yang. 2022.
\newblock Coauthor: Designing a human-ai collaborative writing dataset for exploring language model capabilities.
\newblock In \emph{Proceedings of the 2022 CHI conference on human factors in computing systems}, pages 1--19.

\bibitem[{Li et~al.(2024)Li, Li, Cui, Bi, Wang, Wang, Yang, Shi, and Zhang}]{li2024mage}
Yafu Li, Qintong Li, Leyang Cui, Wei Bi, Zhilin Wang, Longyue Wang, Linyi Yang, Shuming Shi, and Yue Zhang. 2024.
\newblock Mage: Machine-generated text detection in the wild.
\newblock In \emph{Proceedings of the 62nd Annual Meeting of the Association for Computational Linguistics (Volume 1: Long Papers)}, pages 36--53.

\bibitem[{Liang et~al.(2024)Liang, Zhang, Wu, Lepp, Ji, Zhao, Cao, Liu, He, Huang et~al.}]{liang2024mapping}
Weixin Liang, Yaohui Zhang, Zhengxuan Wu, Haley Lepp, Wenlong Ji, Xuandong Zhao, Hancheng Cao, Sheng Liu, Siyu He, Zhi Huang, et~al. 2024.
\newblock Mapping the increasing use of llms in scientific papers.
\newblock \emph{arXiv preprint arXiv:2404.01268}.

\bibitem[{Liu et~al.(2019)Liu, Ott, Goyal, Du, Joshi, Chen, Levy, Lewis, Zettlemoyer, and Stoyanov}]{liu2019roberta}
Yinhan Liu, Myle Ott, Naman Goyal, Jingfei Du, Mandar Joshi, Danqi Chen, Omer Levy, Mike Lewis, Luke Zettlemoyer, and Veselin Stoyanov. 2019.
\newblock Roberta: A robustly optimized bert pretraining approach.
\newblock \emph{arXiv preprint arXiv:1907.11692}.

\bibitem[{Lo et~al.(2020)Lo, Wang, Neumann, Kinney, and Weld}]{lo-wang-2020-s2orc}
Kyle Lo, Lucy~Lu Wang, Mark Neumann, Rodney Kinney, and Daniel Weld. 2020.
\newblock \href {https://doi.org/10.18653/v1/2020.acl-main.447} {{S}2{ORC}: The semantic scholar open research corpus}.
\newblock In \emph{Proceedings of the 58th Annual Meeting of the Association for Computational Linguistics}, pages 4969--4983, Online. Association for Computational Linguistics.

\bibitem[{Lund et~al.(2023)Lund, Wang, Mannuru, Nie, Shimray, and Wang}]{lund2023chatgpt}
Brady~D Lund, Ting Wang, Nishith~Reddy Mannuru, Bing Nie, Somipam Shimray, and Ziang Wang. 2023.
\newblock Chatgpt and a new academic reality: Artificial intelligence-written research papers and the ethics of the large language models in scholarly publishing.
\newblock \emph{Journal of the Association for Information Science and Technology}, 74(5):570--581.

\bibitem[{M~Alshater(2022)}]{m2022exploring}
Muneer M~Alshater. 2022.
\newblock Exploring the role of artificial intelligence in enhancing academic performance: A case study of chatgpt.
\newblock \emph{Available at SSRN}.

\bibitem[{Mitchell et~al.(2023)Mitchell, Lee, Khazatsky, Manning, and Finn}]{mitchell2023detectgpt}
Eric Mitchell, Yoonho Lee, Alexander Khazatsky, Christopher~D Manning, and Chelsea Finn. 2023.
\newblock Detectgpt: Zero-shot machine-generated text detection using probability curvature.
\newblock In \emph{International Conference on Machine Learning}, pages 24950--24962. PMLR.

\bibitem[{Moro et~al.(2001)Moro, Tettamanti, Perani, Donati, Cappa, and Fazio}]{moro2001syntax}
Andrea Moro, Marco Tettamanti, Daniela Perani, Caterina Donati, Stefano~F Cappa, and Ferruccio Fazio. 2001.
\newblock Syntax and the brain: disentangling grammar by selective anomalies.
\newblock \emph{Neuroimage}, 13(1):110--118.

\bibitem[{Nguyen et~al.(2024)Nguyen, Hong, Dang, and Huang}]{nguyen2024human}
Andy Nguyen, Yvonne Hong, Belle Dang, and Xiaoshan Huang. 2024.
\newblock Human-ai collaboration patterns in ai-assisted academic writing.
\newblock \emph{Studies in Higher Education}, pages 1--18.

\bibitem[{Palmer et~al.(2011)Palmer, Gildea, and Xue}]{palmer2011semantic}
Martha Palmer, Daniel Gildea, and Nianwen Xue. 2011.
\newblock \emph{Semantic role labeling}.
\newblock Morgan \& Claypool Publishers.

\bibitem[{Poon and Domingos(2009)}]{poon2009unsupervised}
Hoifung Poon and Pedro Domingos. 2009.
\newblock Unsupervised semantic parsing.
\newblock In \emph{Proceedings of the 2009 conference on empirical methods in natural language processing}, pages 1--10.

\bibitem[{Richburg et~al.(2024)Richburg, Bao, and Carpuat}]{richburg2024automatic}
Aquia Richburg, Calvin Bao, and Marine Carpuat. 2024.
\newblock Automatic authorship analysis in human-ai collaborative writing.
\newblock In \emph{Proceedings of the 2024 Joint International Conference on Computational Linguistics, Language Resources and Evaluation (LREC-COLING 2024)}, pages 1845--1855.

\bibitem[{Salimi and Saheb(2023)}]{salimi2023large}
Ali Salimi and Hady Saheb. 2023.
\newblock Large language models in ophthalmology scientific writing: ethical considerations blurred lines or not at all?
\newblock \emph{American Journal of Ophthalmology}, 254:177--181.

\bibitem[{Solaiman et~al.(2019)Solaiman, Brundage, Clark, Askell, Herbert-Voss, Wu, Radford, Krueger, Kim, Kreps et~al.}]{solaiman2019release}
Irene Solaiman, Miles Brundage, Jack Clark, Amanda Askell, Ariel Herbert-Voss, Jeff Wu, Alec Radford, Gretchen Krueger, Jong~Wook Kim, Sarah Kreps, et~al. 2019.
\newblock Release strategies and the social impacts of language models.
\newblock \emph{arXiv preprint arXiv:1908.09203}.

\bibitem[{Su et~al.(2023)Su, Zhuo, Wang, and Nakov}]{su2023detectllm}
Jinyan Su, Terry~Yue Zhuo, Di~Wang, and Preslav Nakov. 2023.
\newblock Detectllm: Leveraging log rank information for zero-shot detection of machine-generated text.
\newblock \emph{arXiv preprint arXiv:2306.05540}.

\bibitem[{Verma et~al.(2024)Verma, Fleisig, Tomlin, and Klein}]{verma2024ghostbuster}
Vivek Verma, Eve Fleisig, Nicholas Tomlin, and Dan Klein. 2024.
\newblock Ghostbuster: Detecting text ghostwritten by large language models.
\newblock In \emph{Proceedings of the 2024 Conference of the North American Chapter of the Association for Computational Linguistics: Human Language Technologies (Volume 1: Long Papers)}, pages 1702--1717.

\bibitem[{Vigliocco and Vinson(2007)}]{vigliocco2007semantic}
Gabriella Vigliocco and David~P Vinson. 2007.
\newblock Semantic representation.
\newblock \emph{The Oxford handbook of psycholinguistics}, pages 195--215.

\bibitem[{Wang et~al.(2024{\natexlab{a}})Wang, Chen, Jia, Wang, Fang, Wang, Gao, Xie, Xu, Dai et~al.}]{wang2024weaver}
Tiannan Wang, Jiamin Chen, Qingrui Jia, Shuai Wang, Ruoyu Fang, Huilin Wang, Zhaowei Gao, Chunzhao Xie, Chuou Xu, Jihong Dai, et~al. 2024{\natexlab{a}}.
\newblock Weaver: Foundation models for creative writing.
\newblock \emph{arXiv preprint arXiv:2401.17268}.

\bibitem[{Wang et~al.(2024{\natexlab{b}})Wang, Zhou, and Ledo}]{wang2024storyverse}
Yi~Wang, Qian Zhou, and David Ledo. 2024{\natexlab{b}}.
\newblock Storyverse: Towards co-authoring dynamic plot with llm-based character simulation via narrative planning.
\newblock In \emph{Proceedings of the 19th International Conference on the Foundations of Digital Games}, pages 1--4.

\bibitem[{Wang et~al.(2024{\natexlab{c}})Wang, Feng, Hou, Pu, Shen, Liu, Tsvetkov, and He}]{wang2024stumbling}
Yichen Wang, Shangbin Feng, Abe~Bohan Hou, Xiao Pu, Chao Shen, Xiaoming Liu, Yulia Tsvetkov, and Tianxing He. 2024{\natexlab{c}}.
\newblock Stumbling blocks: Stress testing the robustness of machine-generated text detectors under attacks.
\newblock \emph{arXiv preprint arXiv:2402.11638}.

\bibitem[{Wang et~al.(2024{\natexlab{d}})Wang, Mansurov, Ivanov, Su, Shelmanov, Tsvigun, Whitehouse, Afzal, Mahmoud, Sasaki et~al.}]{wang2024m4}
Yuxia Wang, Jonibek Mansurov, Petar Ivanov, Jinyan Su, Artem Shelmanov, Akim Tsvigun, Chenxi Whitehouse, Osama~Mohammed Afzal, Tarek Mahmoud, Toru Sasaki, et~al. 2024{\natexlab{d}}.
\newblock M4: Multi-generator, multi-domain, and multi-lingual black-box machine-generated text detection.
\newblock In \emph{Proceedings of the 18th Conference of the European Chapter of the Association for Computational Linguistics (Volume 1: Long Papers)}, pages 1369--1407.

\bibitem[{Wasi et~al.(2024)Wasi, Islam, and Islam}]{wasi2024llms}
Azmine~Toushik Wasi, Rafia Islam, and Raima Islam. 2024.
\newblock Llms as writing assistants: Exploring perspectives on sense of ownership and reasoning.
\newblock \emph{arXiv preprint arXiv:2404.00027}.

\bibitem[{Wu et~al.(2023)Wu, Yang, Zhan, Yuan, Wong, and Chao}]{wu2023survey}
Junchao Wu, Shu Yang, Runzhe Zhan, Yulin Yuan, Derek~F Wong, and Lidia~S Chao. 2023.
\newblock A survey on llm-generated text detection: Necessity, methods, and future directions.
\newblock \emph{arXiv preprint arXiv:2310.14724}.

\bibitem[{Wu et~al.(2024)Wu, Zhan, Wong, Yang, Yang, Yuan, and Chao}]{wu2024detectrl}
Junchao Wu, Runzhe Zhan, Derek~F Wong, Shu Yang, Xinyi Yang, Yulin Yuan, and Lidia~S Chao. 2024.
\newblock Detectrl: Benchmarking llm-generated text detection in real-world scenarios.
\newblock In \emph{The Thirty-eight Conference on Neural Information Processing Systems Datasets and Benchmarks Track}.

\bibitem[{Xu et~al.(2024)Xu, Wang, An, Liu, and Li}]{xu2024detecting}
Yang Xu, Yu~Wang, Hao An, Zhichen Liu, and Yongyuan Li. 2024.
\newblock Detecting subtle differences between human and model languages using spectrum of relative likelihood.
\newblock In \emph{Proceedings of the 2024 Conference on Empirical Methods in Natural Language Processing}, pages 10108--10121.

\bibitem[{Yan et~al.(2023)Yan, Fauss, Hao, and Cui}]{yan2023detection}
Duanli Yan, Michael Fauss, Jiangang Hao, and Wenju Cui. 2023.
\newblock Detection of ai-generated essays in writing assessment.
\newblock \emph{Psychological Testing and Assessment Modeling}, 65(2):125--144.

\bibitem[{Yang et~al.(2023{\natexlab{a}})Yang, Cheng, Wu, Petzold, Wang, and Chen}]{yang2023dna}
Xianjun Yang, Wei Cheng, Yue Wu, Linda~Ruth Petzold, William~Yang Wang, and Haifeng Chen. 2023{\natexlab{a}}.
\newblock Dna-gpt: Divergent n-gram analysis for training-free detection of gpt-generated text.
\newblock In \emph{The Twelfth International Conference on Learning Representations}.

\bibitem[{Yang et~al.(2023{\natexlab{b}})Yang, Pan, Zhao, Chen, Petzold, Wang, and Cheng}]{yang2023survey}
Xianjun Yang, Liangming Pan, Xuandong Zhao, Haifeng Chen, Linda Petzold, William~Yang Wang, and Wei Cheng. 2023{\natexlab{b}}.
\newblock A survey on detection of llms-generated content.
\newblock \emph{arXiv preprint arXiv:2310.15654}.

\bibitem[{Yuan et~al.(2022)Yuan, Coenen, Reif, and Ippolito}]{yuan2022wordcraft}
Ann Yuan, Andy Coenen, Emily Reif, and Daphne Ippolito. 2022.
\newblock Wordcraft: story writing with large language models.
\newblock In \emph{Proceedings of the 27th International Conference on Intelligent User Interfaces}, pages 841--852.

\bibitem[{Zhang et~al.(2024)Zhang, Gao, Chen, Huang, Huang, Sun, Zhang, Li, Fu, Wan et~al.}]{zhang2024llm}
Qihui Zhang, Chujie Gao, Dongping Chen, Yue Huang, Yixin Huang, Zhenyang Sun, Shilin Zhang, Weiye Li, Zhengyan Fu, Yao Wan, et~al. 2024.
\newblock Llm-as-a-coauthor: Can mixed human-written and machine-generated text be detected?
\newblock In \emph{Findings of the Association for Computational Linguistics: NAACL 2024}, pages 409--436.

\bibitem[{Zhao et~al.(2024{\natexlab{a}})Zhao, Gunn, Christ, Fairoze, Fabrega, Carlini, Garg, Hong, Nasr, Tramer et~al.}]{zhao2024sok}
Xuandong Zhao, Sam Gunn, Miranda Christ, Jaiden Fairoze, Andres Fabrega, Nicholas Carlini, Sanjam Garg, Sanghyun Hong, Milad Nasr, Florian Tramer, et~al. 2024{\natexlab{a}}.
\newblock Sok: Watermarking for ai-generated content.
\newblock \emph{arXiv preprint arXiv:2411.18479}.

\bibitem[{Zhao et~al.(2024{\natexlab{b}})Zhao, Li, and Wang}]{zhao2024permute}
Xuandong Zhao, Lei Li, and Yu-Xiang Wang. 2024{\natexlab{b}}.
\newblock Permute-and-flip: An optimally robust and watermarkable decoder for llms.
\newblock \emph{arXiv preprint arXiv:2402.05864}.

\bibitem[{Zhao et~al.(2023{\natexlab{a}})Zhao, Wang, and Li}]{zhao2023protecting}
Xuandong Zhao, Yu-Xiang Wang, and Lei Li. 2023{\natexlab{a}}.
\newblock Protecting language generation models via invisible watermarking.
\newblock In \emph{International Conference on Machine Learning}, pages 42187--42199. PMLR.

\bibitem[{Zhao et~al.(2023{\natexlab{b}})Zhao, Song, Duah, Macbeth, Carter, Van, Bravo, Klenk, Sick, and Filipowicz}]{zhao2023more}
Zoie Zhao, Sophie Song, Bridget Duah, Jamie Macbeth, Scott Carter, Monica~P Van, Nayeli~Suseth Bravo, Matthew Klenk, Kate Sick, and Alexandre~LS Filipowicz. 2023{\natexlab{b}}.
\newblock More human than human: Llm-generated narratives outperform human-llm interleaved narratives.
\newblock In \emph{Proceedings of the 15th Conference on Creativity and Cognition}, pages 368--370.

\end{thebibliography}

\newpage
\appendix

\section{Benchmark Dataset}

HART will be released under Creative Commons license, which is also the license publicly available by all the source data.

\subsection{Domains and Languages}
\label{app:datasets}

HART encompasses four domains and five languages, with each language featuring 2,000 development samples and 2,000 test samples. Within these datasets, the samples are evenly distributed across four types (0, 1, 2, and 3). For every domain, human-written texts (type 0) are collected from specific sources, while AI-generated texts (type 3) are produced using prompts outlined in Table \ref{tab:prompt_data_generation}.

\paragraph{Student Essay.}  
We randomly select 1,000 essays from the Automated Student Assessment Prize (ASAP) 2.0 \cite{scott2024asap}, each accompanied by a title and a prompt. These prompts are utilized to prompt LLMs to generate corresponding essays. Additionally, metadata such as `{\it race ethnicity}', `{\it gender}', and `{\it grade level}' is recorded for potential future analyses.

\paragraph{ArXiv Intro.}  
To build this dataset, we collect 1,000 computer science papers from arXiv \cite{arxiv2024} by crawling PDFs published between 2020 and 2024, randomly selecting 200 papers per year. Using S2ORC \cite{lo-wang-2020-s2orc}, the PDFs are parsed to extract titles and introductions. These titles are then used to prompt LLMs to generate new paper introductions. The inclusion of publication year also provides a basis for analyzing distribution shifts over time.

\paragraph{Creative Writing.}  
We randomly pull 1,000 samples from WritingPrompts \cite{fan2018hierarchical}, with each sample paired with a corresponding prompt. These prompts serve as triggers for LLMs to create new fictional stories.

\paragraph{CC News.} 
For this dataset, we gather 1,000 news articles in each of five languages -- English, Chinese, French, Spanish, and Arabic -- sourced from Common Crawl \cite{Hamborg2017}. The news headlines are used to prompt LLMs to generate full news articles.

\begin{table}[t]
    \centering\small
    \begin{NiceTabular}{p{0.95\linewidth}}
        \toprule
        \textbf{Student Essay:} {\it Write a student essay (no title) in \{n\_words\} words (split into \{n\_paragraphs\} paragraphs) based on the given \{field\}:$\backslash$n \{field\_value\}} \\
        \addlinespace[0.5em]
        \textbf{ArXiv Intro:} {\it Write an introductory section (no section name) for an academic paper in \{n\_words\} words (split into \{n\_paragraphs\} paragraphs) based on the given \{field\}:$\backslash$n \{field\_value\}} \\
        \addlinespace[0.5em]
        \textbf{Creative Writing:} {\it Write a creative story (no title) in \{n\_words\} words (split into \{n\_paragraphs\} paragraphs) based on the given \{field\}:$\backslash$n \{field\_value\}} \\
        \addlinespace[0.5em]
        \textbf{CC News:} {\it Write a news article (no title) in \{n\_words\} words (split into \{n\_paragraphs\} paragraphs) based on the given \{field\}:$\backslash$n \{field\_value\}} \\
        \addlinespace[0.5em]
        \textbf{Multi-lingual CC News:} {\it Write a news article (no title)  in \{lang\} language in \{n\_words\} words (split into \{n\_paragraphs\} paragraphs) based on the given \{field\}:$\backslash$n \{field\_value\}} \\
        \bottomrule
    \end{NiceTabular}
    \caption{Prompts for data generation, where the {\it field} could be either `{\it title}' or `{\it prompt}' depending on their availability for each data source.}
    \label{tab:prompt_data_generation}
\end{table}

\subsection{Automatic Refinement}
\label{app:refinement}

We use the following prompts for automatic refinement of human-written texts.

\paragraph{Prompt for Polishing:} ``{\it Polish the text to enhance its readability, coherence, and flow. Correct any grammatical, punctuation, and style errors, but ensure the core content remains unchanged:$\backslash$n\{generation\}}''

\paragraph{Prompt for Restructuring:} ``{\it Restructure the text to improve its logical flow and coherence by rearranging paragraphs, sections, or sentences for enhanced clarity and fluency:$\backslash$n\{generation\}}''

\subsection{Humanizing}

\subsubsection{Automatic Humanizing}
\label{app:humanizing}

We use the following prompts for humanizing AI-generated texts automatically.

\paragraph{Prompt for Diversifying:} ``{\it Revise the text to enrich its language diversity, employing varied sentence structures, synonyms, and stylistic nuances, while preserving the original meaning:$\backslash$n\{generation\}}''

\paragraph{Prompt for Mimicking:} ``{\it Rewrite the text using the same language style, tone, and expression as the reference text. Focus on capturing the unique vocabulary, sentence structure, and stylistic elements evident in the reference:$\backslash$n\{generation\}$\backslash$n$\backslash$n\# Reference Text:$\backslash$n\{reference\}}''

\begin{table}[t]
    \centering\small
    \begin{tabular}{llc}
        \toprule
        \bf AI Tool & \bf URL & \bf Used \\
        \midrule
        BypassGPT & https://bypassgpt.ai/ & Y \\
        Humbot & https://humbot.ai/ & Y \\
        Undetectable AI & https://undetectable.ai/ & Y \\
        Semihuman AI & https://semihuman.ai/ \\
        HIX Bypass & https://bypass.hix.ai/ \\
        AI Humanizer & https://aihumanizer.ai/ \\
        StealthGPT & https://stealthgpt.ai/ \\
        GPTinf & https://stealthgpt.ai/ \\
        WriteHuman & https://writehuman.ai/ \\
        StealthWriter & https://rewritify.ai/ \\
        Phrasly LLC & https://phrasly.ai/ \\
        HIX.AI & https://bypass.hix.ai \\
        AISEO Humanizer & https://aiseo.ai/ \\
        Humanize AI Pro & https://www.humanizeai.pro/ \\
        Smodin & https://smodin.io/ \\
        Rewritify & https://www.rewritify.ai \\
        \bottomrule
    \end{tabular}
    \caption{Humanizing tools that bypass detectors.}
    \label{tab:list_humanizing_system}
\end{table}

\subsubsection{External Humanizing Tools}
\label{app:external_tools}

There are various AI humanizing tools that are developed to bypass detectors. We list a few in Table \ref{tab:list_humanizing_system}, where the first three are used to produce our type-2 texts. We demonstrate that these tools all alter texts at the surface level, where the content feature has strong resilience.

\subsubsection{Human Editing}
\label{app:human_editing}

We employ five annotators from a specialized annotation company, consisting of three individuals with professional expertise in English and two with a background in computer science. The team includes three women and two men, aged between 22 and 41, all of whom have Chinese as their native language. Each annotator is responsible for revising 50 AI-generated texts, resulting in a total of 250 human-edited samples. 

The editing process is carried out at three levels: word, sentence, and paragraph. At the word level, synonyms are used to replace existing words; at the sentence level, syntax alterations are made; and at the paragraph level, the logical flow of sentences is reorganized. Annotators are asked to apply these three types of edits in equal proportion, ensuring that over 50\% of the original content is modified, as described in Table \ref{tab:human_editing_instruction}. Additionally, a separate annotator reviews 10\% of the texts to verify that the edits preserve the original meaning while ensuring that the revised texts remain fluent and comprehensible.

\begin{table*}[t]
    \centering\small
    \begin{tabular}{lcccccc}
        \toprule
        \bf Detector & \bf English & \bf Chinese & \bf French & \bf Spanish & \bf Arabic & \bf ALL (TPR5\%) \\
        \midrule
        \multicolumn{7}{c}{\bf \colorbox{red!60}{Level-3} Detection Task} \\
        \hdashline
        Log-Perplexity & 0.7370 & 0.8599 & 0.8471 & 0.8668 & 0.6290 & 0.7489 (26\%) \\
        Log-Rank & 0.7665 & \bf 0.8701 & 0.8645 & \bf 0.8770 & \bf 0.6327 & 0.7647 (25\%) \\
        LRR & 0.8466 & 0.8625 & 0.8706 & 0.8744 & 0.6117 & 0.7651 (21\%) \\
        \hdashline
        Fast-Detect & 0.8551 & 0.8655 & 0.8662 & 0.8310 & 0.5871 & 0.8118 (48\%) \\
        $C_2$ (Fast-Detect) & 0.7084 & 0.7121 & 0.7404 & 0.7163 & 0.5657 & 0.6910 (18\%) \\
        $C_2$-$T$ (Fast-Detect) & 0.8600 & 0.8459 & 0.8538 & 0.8397 & 0.5879 & 0.8065 (48\%) \\
        \hdashline
        Binoculars & \bf 0.8698 & 0.8698 & 0.8814 & 0.8474 & 0.5754 & 0.7990 (48\%) \\
        $C_2$ (Binoculars) & 0.7177 & 0.7117 & 0.7633 & 0.7318 & 0.5507 & 0.6882 (19\%) \\
        $C_2$-$T$ (Binoculars) & 0.8698 & 0.8495 & 0.8587 & 0.8548 & 0.5476 & 0.7924 (49\%) \\
        \hdashline
        Glimpse & 0.8310 & 0.8868 & 0.8793 & 0.8382 & 0.7950 & 0.8323 (51\%) \\
        $C_2$ (Glimpse) & 0.7422 & 0.7182 & 0.7434 & 0.7382 & 0.6952 & 0.6958 (13\%) \\
        $C_2$-$T$ (Glimpse) & 0.8257 & 0.8681 & \bf 0.8853 & 0.8729 & \bf 0.8031 & \bf 0.8481 (53\%) \\
        \midrule
        \multicolumn{7}{c}{\bf \colorbox{red!40}{Level-2} Detection Task} \\
        \hdashline
        Log-Perplexity & 0.3909 & 0.8660 & 0.6955 & 0.7963 & 0.4679 & 0.6287 (14\%) \\
        Log-Rank & 0.4130 & \bf 0.8769 & 0.7066 & 0.7959 & 0.4686 & 0.6350 (12\%) \\
        LRR & 0.5296 & 0.8748 & 0.7118 & 0.7644 & 0.4777 & 0.6380 (11\%) \\
        \hdashline
        Fast-Detect & 0.6665 & 0.8361 & 0.7728 & 0.6961 & 0.4658 & 0.7007 (37\%) \\
        $C_2$ (Fast-Detect) & 0.7412 & 0.7778 & 0.7819 & 0.7454 & \bf 0.6099 & 0.7336 (28\%) \\
        $C_2$-$T$ (Fast-Detect) & 0.8242 & 0.8295 & 0.8344 & 0.7837 & 0.5867 & \bf 0.7793 (42\%) \\
        \hdashline
        Binoculars & 0.6770 & 0.8383 & 0.7779 & 0.7115 & 0.4543 & 0.6929 (37\%) \\
        $C_2$ (Binoculars) & 0.7492 & 0.7803 & 0.7968 & 0.7587 & 0.5955 & 0.7265 (28\%) \\
        $C_2$-$T$ (Binoculars) & \bf 0.8310 & 0.8234 & \bf 0.8464 & 0.7955 & 0.4978 & 0.7515 (38\%) \\
        \hdashline
        Glimpse & 0.5953 & 0.8123 & 0.7511 & 0.7269 & 0.6813 & 0.6921 (33\%) \\
        $C_2$ (Glimpse) & 0.6990 & 0.7569 & 0.7444 & 0.7110 & 0.7059 & 0.6860 (14\%) \\
        $C_2$-$T$ (Glimpse) & 0.7094 & 0.8258 & 0.8257 & \bf 0.8083 & \bf 0.7969 & 0.7776 (41\%) \\
        \midrule
        \multicolumn{7}{c}{\bf \colorbox{yellow!60}{Level-1} Detection Task} \\
        \hdashline
        Log-Perplexity & 0.3960 & 0.8483 & 0.6629 & \bf 0.7890 & 0.4876 & 0.6154 (08\%) \\
        Log-Rank & 0.4042 & \bf 0.8519 & 0.6640 & 0.7824 & 0.4830 & 0.6140 (07\%) \\
        LRR & 0.5009 & 0.8309 & 0.6480 & 0.7288 & 0.4760 & 0.6070 (08\%) \\
        Glimpse & \\
        \hdashline
        Fast-Detect & 0.6897 & 0.8349 & 0.7510 & 0.7331 & 0.4359 & 0.7032 (30\%) \\
        $C_2$ (Fast-Detect) & 0.7021 & 0.7237 & 0.6963 & 0.7105 & 0.5678 & 0.6820 (22\%) \\
        $C_2$-$T$ (Fast-Detect) & 0.7770 & 0.7997 & \bf 0.7749 & 0.7669 & 0.4798 & 0.7288 (32\%) \\
        \hdashline
        Binoculars & 0.6969 & 0.8394 & 0.7484 & 0.7461 & 0.4286 & 0.7053 ({\bf 33\%}) \\
        $C_2$ (Binoculars) & 0.6959 & 0.7234 & 0.7042 & 0.7137 & 0.5521 & 0.6752 (21\%) \\
        $C_2$-$T$ (Binoculars) & \bf 0.7843 & 0.8041 & 0.7637 & 0.7657 & 0.4639 & 0.7264 (33\%) \\
        \hdashline
        Glimpse & 0.5600 & 0.7928 & 0.6933 & 0.7034 & 0.6673 & 0.6596 (24\%) \\
        $C_2$ (Glimpse) & 0.5815 & 0.6481 & 0.6456 & 0.6192 & 0.6052 & 0.6003 (10\%) \\
        $C_2$-$T$ (Glimpse) & 0.6386 & 0.7904 & 0.7336 & 0.7634 & \bf 0.7638 & {\bf 
 0.7302} (24\%) \\
        \bottomrule
    \end{tabular}
    \caption{Results in CC News of HART, covering five languages. The best AUROC and TPR5\% are marked in \textbf{bold}. The column `{\it ALL}' denotes a mixture of languages.}
    \label{tab:results_multilingual}
\end{table*}

\subsubsection{Data Coverage}
\label{app:data_coverage}
HART encompasses four domains and five languages as Table \ref{tab:datasets}, which are generated by six LLMs and four decoding parameters. Specifically, the dataset leverages six language models -- gpt-3.5-turbo, gpt-4o, claude-3.5-sonnet, gemini-1.5-pro, llama-3.3-70b-instruct, and qwen-2.5-72b-instruct -- to generate data, with a random model selected for each sample. As for decoding parameters, a temperature is randomly chosen from the range $[0.8, 1.0, 1.2]$, a top-$p$ from $[0.96, 1.0]$, and both frequency and presence penalties from the range $[0.0, 1.0]$ for each sample.

\begin{figure*}[t]
    \centering
    \includegraphics[trim={60pt 20pt 70pt 2pt},clip,width=1.0\linewidth]{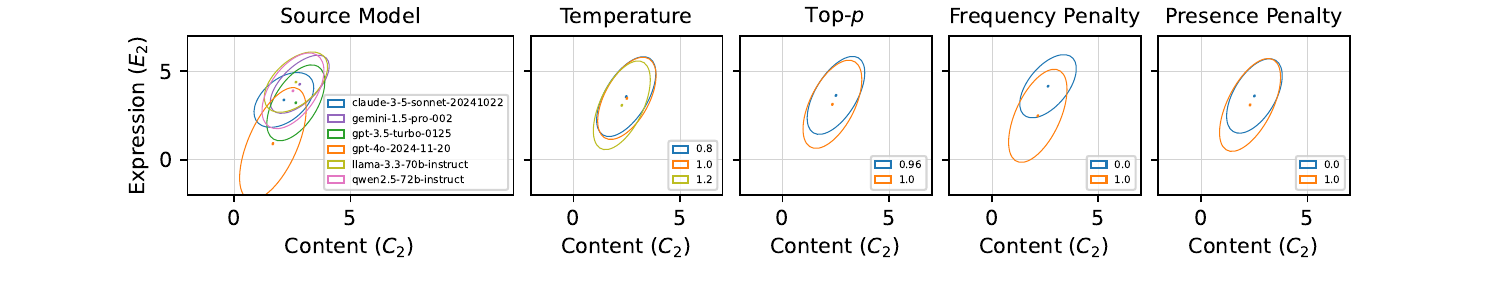}   
    \caption{The impact of source model and decoding parameters to generated texts.}
    \label{fig:group_2d_distrib}
\end{figure*}

\begin{figure*}[t]
    \centering
    \includegraphics[trim={0pt 0pt 0pt 0pt},clip,width=0.7\linewidth]{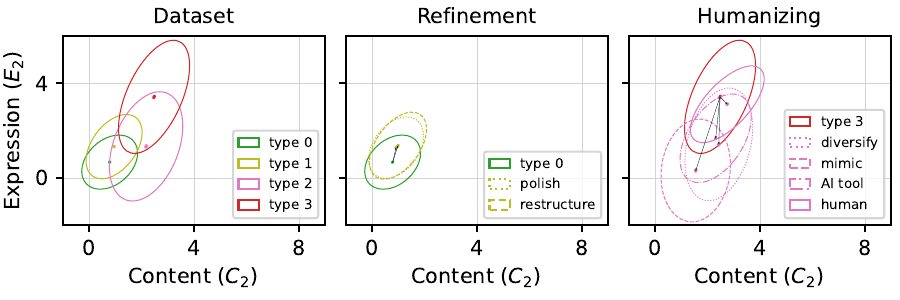}   
    \caption{2D distribution of texts using conditional probability curvature as a metric. The center points represent the means while ellipses the standard deviations.}
    \label{fig:2d_distribution}
\end{figure*}

\section{Baseline Detectors}
\label{app:detectors}

\paragraph{RoBERTa (ChatGPT)}
\cite{guo2023hc3} refers to a RoBERTa-base model \cite{liu2019roberta} that has been fine-tuned on the HC3 \cite{guo2023hc3} dataset. This dataset includes responses written by humans and ChatGPT across a variety of fields such as Reddit, medicine, finance, and law. We use this model as a representative baseline for trained detectors. \footnote{https://huggingface.co/Hello-SimpleAI/chatgpt-detector-roberta} 

\paragraph{RADAR}
\cite{hu2023radar} is trained on Vicuna-7B, employing a generative adversarial framework. In this setup, a paraphraser is optimized to deceive the detector, while the detector itself learns to recognize outputs generated by the paraphraser. \footnote{https://huggingface.co/TrustSafeAI/RADAR-Vicuna-7B}

\paragraph{Log-Perplexity and Log-Rank}
\cite{gehrmann2019gltr,solaiman2019release,ippolito2020automatic} are simple yet effective baseline methods. Log-perplexity measures the logarithmic perplexity of a scoring model, while Log-rank computes the average logarithm of token ranks in descending probability order. For this study, we use falcon-7B as the scoring model, which has shown superior performance compared to smaller models.

\paragraph{LRR}
\cite{su2023detectllm} is a detector based on perplexity, calculated by dividing the perplexity by the log-rank of a scoring model. Similar to others, falcon-7B serves as the scoring model in our experiments.

\paragraph{Fast-Detect}
\cite{bao2024fast} employs a perplexity-based detection approach. It calculates a metric named conditional probability curvature by subtracting perplexity on a scoring model from the cross-entropy between the scoring model and a sampling model. The original implementation uses gpt-j-6B as the sampling model and gpt-neo-2.7B as the scoring model. However, we observe that using a larger model considerably improves detection. To ensure fair comparisons, we use falcon-7B-instruct as the scoring model and falcon-7B as the sampling model, similar to Binoculars.

\paragraph{Binoculars}
\cite{hans2024spotting} is another perplexity-based detection method, which operates by dividing the perplexity of a scoring model (referred to as the performer) by the cross-entropy between the performer model and an observer model. In our experiments, we adhere to the original setup, where falcon-7B-instruct acts as the performer and falcon-7B as the observer.

\paragraph{Glimpse}
\cite{bao2025glimpse} is a variation of Fast-Detect that utilizes the proprietary gpt-3.5-turbo-0301 model. It approximates the full token probability distribution using a partial observation retrieved from the Completion API.

\section{Results and Analysis}

\subsection{Results on Multiple Languages}
\label{app:multilingual}

As shown in Table \ref{tab:results_multilingual}, 2D detectors demonstrate clear superiority over baseline models in level-2 and level-1 detection tasks across five languages, highlighting the effectiveness of the 2D framework across all tested languages. Among the existing detectors, Glimpse, powered by gpt-3.5-turbo-0301, outperforms both Fast-Detect and Binoculars, which are based on falcon-7B and falcon-7B-instruct. We hypothesize that this advantage stems from the stronger multilingual capabilities of gpt-3.5-turbo.

\subsection{Analysis on Data Distribution}

\subsubsection{Impact of Parameters}
\label{app:analysis_params}

As illustrated in Figure \ref{fig:group_2d_distrib}, each factor uniquely affects the distribution of generated texts.

\paragraph{Source Model.}  
The source model plays the most pivotal role in shaping the distribution. Among the six models, gpt-4o stands out by generating the most diverse texts across both expression and content dimensions.

\paragraph{Temperature.}  
In general, increasing the temperature leads to greater diversity in the generated texts. However, the variations in diversity are relatively minor.

\paragraph{Top-$p$.}  
Similar to temperature, larger values of $p$ result in more diverse outputs, but the overall differences remain limited.

\paragraph{Frequency Penalty.}  
Frequency penalty during decoding significantly influences the distribution of generated texts, with higher penalties tending to produce more human-like outputs.

\paragraph{Presence Penalty.}  
Compared to frequency penalty, presence penalty has a smaller impact. Nonetheless, higher penalties generally result in more human-like text generation.

\subsubsection{Impact of Refinement and Humanizing}
\label{app:analysis_refinement_humanizing}
As illustrated in Figure \ref{fig:2d_distribution}, both refinement and humanizing introduce significant changes to the distribution.

\paragraph{Refinement.}  
The process of refining, which involves polishing and restructuring, shifts the distribution upward, indicating a notable influence along the expression dimension. However, the variations between the two refinement techniques are relatively minimal.

\paragraph{Humanizing.}  
Various humanizing techniques affect the distribution differently. Human editing induces the smallest changes, maintaining a center close to the origin. In contrast, AI tools produce a more pronounced impact, though the shift predominantly occurs along the expression dimension. The ``Diversify'' technique yields results similar to external AI tools, while ``Mimic'' causes the most substantial distribution shift.

\subsection{Discussion on Alternative Solutions}

Due to their straightforward nature, the prompting techniques serve as suitable options for a proof-of-concept. However, there are numerous alternative methods for representing content, including structural representations like abstract meaning representations \cite{banarescu2013abstract}, semantic role labeling \cite{palmer2011semantic}, semantic parsing \cite{poon2009unsupervised}, knowledge graphs \cite{hogan2021knowledge}, and frame semantics \cite{fillmore2006frame}, as well as neural representations \cite{vigliocco2007semantic}. For instance, neural representations might be a more effective approach since they bypass the need to depend on external large language models (LLMs) for extracting content features. Further exploration of these possibilities is deferred to future work.

\begin{table*}[t]
    \centering\small
    \begin{NiceTabular}{p{0.95\linewidth}}
    \toprule
    \textbf{Instruction for Manual Editing of AI-Generated Texts}
    
    The goal of this task is to manually edit AI-generated essays, paper introductions, creative writings, and news articles to improve their language quality while retaining the original intended meaning. Follow the steps and guidelines below carefully to ensure consistency and quality.
    
    ---
    
    \textbf{General Guidelines}
    
    1. \textbf{Preserve Original Meaning}: 
    
       - Your edits must not alter the intended meaning or factual content of the original text. Focus solely on improving language clarity, expression, and flow.
    
    2. \textbf{Balance of Edits}:
    Ensure that your edits improve the language across three levels: word, sentence, and paragraph. Distribute your edits so that:
    
         - Word-level modifications account for about 1/3 of your changes.
         
         - Sentence-level modifications account for about 1/3 of your changes.
         
         - Paragraph-level modifications account for about 1/3 of your changes.
    
    3. \textbf{Volume of Edits}:
    
       - The cumulative changes you make should amount to editing more than half of the total word count of the text. Be thorough in your revisions.
    
    ---
    
    \textbf{Types of Edits}
    
    \textbf{1. Word-Level Editing}
    
       - Replace repetitive or vague words with more precise synonyms.
       
       - Improve word choice to match the tone and style of the piece (e.g., academic, formal, journalistic, creative).
       
       - Correct incorrect usage of words, awkward phrasing, or redundant expressions.
    
       \textbf{Example}:
       
       Original: "The results were really very significant."
       
       Edited: "The results were highly significant."
    
    \textbf{2. Sentence-Level Editing}
    
       - Adjust sentence structures to enhance readability and fluency. This includes:
       
         - Breaking down long, convoluted sentences into shorter, clear ones.
         
         - Combining choppy or fragmented sentences for better flow.
         
         - Reorganizing sentence components for coherence and logic.
         
       - Fix issues with grammar, punctuation, and syntax where necessary.
       
       - Ensure variety in sentence structure to avoid monotony.
    
       \textbf{Example}:
       
       Original: "The team successfully completed the project, which was a very crucial step in their plan, and they presented it to stakeholders two days later."
       
       Edited: "The team successfully completed this crucial step in their plan and presented the project to stakeholders two days later."
    
    \textbf{3. Paragraph-Level Editing}
    
       - Rearrange sentences within the paragraph to improve logical progression and argument clarity.
       
       - Merge or split paragraphs when necessary for better organization or flow.
       
       - Add transitional phrases if needed to improve coherence between sentences and paragraphs.
       
       - Ensure that the paragraph aligns with the overall tone and intent of the text.
    
       \textbf{Example}:
       
       Original: ``Climate change is a growing concern worldwide. The effects of climate change include rising temperatures, melting polar ice, and severe weather events. Many governments are implementing policies to mitigate these effects. Public awareness around climate change has also been increasing over recent years. Organizations are focusing on educating individuals and communities about sustainable practices.''
    
       Edited:  ``Climate change is an increasingly urgent issue with global implications. Its effects, such as rising temperatures, melting polar ice, and severe weather events, are becoming more evident. In response, many governments are enacting policies to address these challenges. At the same time, public awareness has grown significantly, driven by organizations that educate communities about sustainable practices.''
    
    ---
    
    \textbf{Step-by-Step Workflow}
    
    1. \textbf{Understand the Text}:
    
       - Read the entire text carefully to grasp its main ideas, tone, and intent before making any changes.
    
    2. \textbf{Edit with Balance and Intent}:
    
       - As you edit, keep track of the types of changes you are making (word-level, sentence-level, paragraph-level) and ensure an even distribution across the three levels.
       
       - Avoid over-editing in one specific area (e.g., only doing word-level tweaks).
    
    3. \textbf{Meet the Edit Requirement}:
    
       - Ensure that more than 50\% of the text has been edited after revisions. Track your changes to confirm this.
    
    4. \textbf{Review and Finalize}:
    
       - Re-read your edited version to confirm it retains the original meaning and intention.
       
       - Check that the language is smooth, natural, and appropriate for the target audience and genre.
    
    \\
    \bottomrule
    \end{NiceTabular}
    \caption{Instruction for human editing.}
    \label{tab:human_editing_instruction}
\end{table*}

\end{document}